\documentclass[runningheads]{llncs}

\usepackage{eccv}

\usepackage{eccvabbrv}

\usepackage{graphicx}
\usepackage{booktabs}
\usepackage{amsmath}
\usepackage{subfiles}

\usepackage{algorithmic}
\usepackage[linesnumbered,ruled,vlined]{algorithm2e}

\definecolor{mygray}{gray}{0.8}

\SetKwInput{KwInput}{Input}  
\SetKwInput{KwOutput}{Output} 

\usepackage[accsupp]{axessibility}  

\usepackage{hyperref}

\usepackage{orcidlink}

\usepackage[dvipsnames]{xcolor}

\usepackage{adjustbox}
\usepackage{multirow}
\usepackage{colortbl}
\usepackage{xcolor}
\usepackage{wrapfig}
\usepackage{graphicx}
\usepackage{caption}
\usepackage{subcaption}
\definecolor{mygray}{gray}{0.8}

\begin{document}

\title{BurstGP: Enhancing Raw Burst Image Super Resolution with Generative Priors} 

\titlerunning{BurstGP}


\author{Dong Huo\inst{1*} \and
Tristan Aumentado-Armstrong\inst{1*} \and
Samrudhdhi B. Rangrej\inst{1*} \and
Maitreya Suin\inst{1} \and
Angela Ning Ye\inst{1} \and
Zhiming Hu\inst{1} \and
Amanpreet Walia\inst{1} \and
Amirhossein Kazerouni\inst{1,2,3,4} \and
Konstantinos G. Derpanis\inst{1,2,3,5}
Iqbal Mohomed\inst{1} \and
Alex Levinshtein\inst{1}
}

\authorrunning{D.~Huo et al.}

\institute{\mbox{AI Center - Toronto, Samsung Electronics \and 
University of Toronto} \and 
\mbox{Vector Institute \and 
University Health Network \and 
York University} \\
\textsuperscript{*} Equal contribution}

\onecolumn{
 \maketitle
 \centerline{
\includegraphics[width=0.99\textwidth]{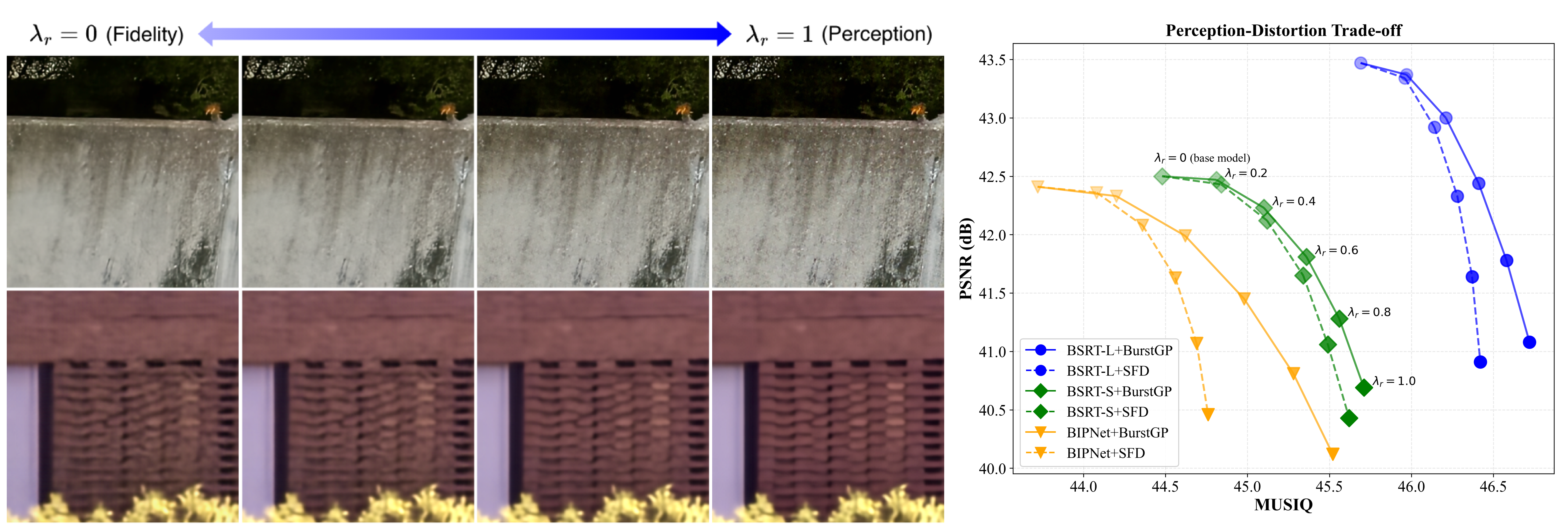}
\label{fig:teaser}

}
\captionof{figure}{
\textbf{Controlling the perception–distortion trade-off with BurstGP.}
Building on a video diffusion model, our method outputs a linear RGB correction to a non-diffusion burst image super-resolution (BISR) output, controlled by a parameter, $\lambda_r \in [0,1]$. When $\lambda_r = 0$, the base model output is unchanged, while increasing $\lambda_r$ progressively enhances perceptual details. 
\textbf{Left}: qualitative results from BurstGP with BSRT-L~\cite{luo2022bsrt}, from high fidelity (low $\lambda_r$) to high perceptual quality (high $\lambda_r$).
\textbf{Right:} Traversing the perception–distortion trade-off (PSNR vs.\ MUSIQ) across several state-of-the-art BISR base models.
Across $\lambda_r$ values,
our multiframe-aware BurstGP offers a superior realism-fidelity tradeoff compared to single-frame diffusion (SFD).
}
\label{fig:teaser}
}

\begin{abstract}
  Burst image super resolution (BISR) aims to construct a single high-resolution (HR) image by aggregating information from multiple low-resolution (LR) frames, relying on temporal redundancy and spatial coherence across the burst. While conventional methods achieve impressive results, they often struggle with complex textures and oversmoothing. 
  Diffusion models, particularly those pretrained on high-quality data, have shown remarkable capability in generating realistic details for image and video super-resolution. However, their potential remains largely under-explored in BISR, where existing approaches typically rely on task-specific diffusion models trained from scratch and operate on single-frame reconstructions.
  In this work, we propose \textbf{BurstGP}, a novel diffusion-based solution for BISR, which leverages generative priors of recent foundation models to overcome these issues. In particular, we build a multiframe-aware diffusion model on top of a conventional BISR approach, which boosts image quality with minimal loss to fidelity. 
  Further, we introduce (i) a novel degradation-aware conditioning mechanism, which controls synthesis of fine details based on the estimated degradation in the input, and (ii) a robust sRGB-to-lRGB inverter, enabling us to utilize generative multiframe (video) sRGB priors, while operating with raw input and lRGB output images. Empirically, we demonstrate that BurstGP outperforms the existing state of the art, both quantitatively (especially with respect to perceptual metrics, including MUSIQ and LPIPS) and qualitatively. In particular, our proposed method excels at recovering richer textures and finer structural details, highlighting the potential of video priors for BISR over traditional methods.
\end{abstract}

\section{Introduction}
\label{sec:intro}

Burst image super-resolution (BISR) reconstructs a single high-resolution (HR) image from a sequence of low-resolution (LR) observations by exploiting complementary information present within and across multiple frames.  By aggregating information across 
the burst, BISR can overcome limitations inherent to single-image super resolution (SR).
While traditional BISR methods have evolved significantly from early interpolation-based techniques~\cite{dai2007soft, li2001new} to sophisticated deep learning approaches~\cite{bhat2021deep, dudhane2023burstormer, kang2024burstm, bhat2021deep2, dudhane2022burst, luo2022bsrt, wei2023towards, luo2021ebsr}, several fundamental challenges persist in achieving optimal reconstruction quality.

The conventional BISR pipeline typically involves three key steps: (i) motion estimation and compensation to align frames, either through explicit optical flow estimation~\cite{kang2024burstm, bhat2021deep} or implicit deformable convolution~\cite{dudhane2023burstormer, dudhane2022burst}, (ii) fusion of aligned information, and (iii) reconstruction of the HR image. Recent methods adopt deep neural networks, particularly convolutional neural networks (CNNs) \cite{LeCunBBH98} and vision transformers (ViTs) \cite{DosovitskiyB0WZ21}, to learn complex mappings between LR inputs and HR outputs. However, these methods often struggle with several critical aspects: maintaining structural consistency while enhancing fine details, handling complex motion patterns and occlusions, and effectively suppressing noise and compression artifacts that vary across input frames.

Diffusion models~\cite{ho2020denoising, rombach2022high, peebles2023scalable}
have advanced generative image processing, enabling high-fidelity image synthesis and restoration.
These models operate through an iterative denoising process, gradually refining a noisy input into a high-quality output. Recently, diffusion models have shown remarkable success in single-image SR~\cite{lin2024diffbir, wang2024exploiting, sun2025pixel} and video SR~\cite{wang2025seedvr, xie2025star, chen2025dove}, but their application to multi-frame raw burst image SR remains largely unexplored. This represents a significant gap in the literature, as the unique properties of video diffusion models, especially those pre-trained on large high-quality datasets, could potentially address many of the limitations inherent in current BISR approaches.

In this work, we present BurstGP, the first comprehensive framework for applying diffusion-based video SR priors to the task of raw-space BISR, which takes advantage of pre-trained generative priors on high-quality sRGB images without suffering from a domain gap. Specifically, inspired by DiffBIR~\cite{lin2024diffbir} which enhances conventional super-resolution models using an image diffusion, the outputs of conventional BISR methods~\cite{kang2024burstm, dudhane2023burstormer,wei2023towards} are first fed to a differentiable image signal processing (ISP) pipeline to match the domain of the diffusion model. The diffusion model then refines the reconstruction in a degradation-aware fashion to enhance fine details. Finally, a robust ISP inversion is proposed to convert the sRGB results back to camera sensor space (linear RGB space). The proposed method offers several key advantages over existing approaches: 
\begin{itemize}
    \item Building on video restoration architectures, we provide a mechanism for \textit{multi}-image (burst) restoration within a generative diffusion model that leverages strong spatiotemporal priors, improving performance over applying diffusion to \textit{single}-image outputs of burst methods (see Fig. \ref{fig:teaser}).
    \item 
        We devise a degradation-aware conditioning mechanism, which enables the model to dynamically adjust the timestep embedding, and further controls the level of details synthesis.
    \item
    We propose an efficient and robust inverse ISP pipeline that enables conversion between sRGB (where our video prior is defined) and linear RGB, and avoids the instability of naive inversion.
%
    
  
\end{itemize}
Empirically, our proposed method, BurstGP, significantly improves perceptual quality over conventional BISR approaches. 
    In addition, it enables intuitive control over the model's perception-distortion tradeoff, as shown in Fig. \ref{fig:teaser}.

\section{Related Work}
\label{sec:relwork}
\textbf{BISR.}
Traditional approaches for BISR frame the task as inverse imaging, where multiple raw LR observations are related to an unknown linear HR image through subpixel shifts (motion), blur, downsampling, noise, and mosaicing. Classical methods therefore rely on explicit registration, followed by robust multi-frame fusion to recover high-frequency details~\cite{irani1991improving, hasinoff2016burst}. 
%
Recently, deep learning has substantially advanced burst image restoration, with early works combining classical formulations with new architectures~\cite{mildenhall2018burst,bhat2021deep2}.
For benchmarking, BurstSR~\cite{bhat2021deep} provides \textit{real-world} data, albeit from different devices (incurring misalignment), while RealBSR~\cite{wei2023towards} improves misalignment via optical zoom on the same device.
As real paired data is difficult to collect, 
Bhat \etal\cite{bhat2023self} leverage self-supervised learning, while others 
use optical-flow guided warping~\cite{kang2024burstm, bhat2021deep} or implicitly align via deformable convolutions~\cite{dudhane2023burstormer, dudhane2022burst, luo2022bsrt}
to handle misalignment. 
More recent models investigate long-range dependencies and more efficient aggregation~\cite{di2025qmambabsr}.
BSRD~\cite{tokoro2024burst} incorporates diffusion into BISR by conditioning the diffusion process on aligned burst features and refining an initial deterministic single-frame SR estimation through diffusion sampling. However, their model is trained specifically for burst SR and does not leverage large pretrained generative priors. In contrast to methods that intertwine fusion and generation early, our design explicitly separates fidelity-critical multi-frame reconstruction from perceptual refinement. BurstGP exploits a pretrained diffusion prior learned from high-quality sRGB data. Through an ISP-based domain bridge and degradation-aware conditioning, BurstGP integrates this prior into raw-space BISR and enables explicit control over the perception-distortion trade-off.

\noindent
\textbf{Diffusion-based SR: Images. }
State-of-the-art generative models, often based on diffusion \cite{ho2020denoising} or flow matching \cite{lipmanflow}, can construct images of unprecedented quality \cite{podell2023sdxl,labs2025flux,baldridge2024imagen}. 
Beyond 
generation and editing \cite{huang2025diffusion}, such models can also serve as powerful priors for SR \cite{wang2024exploiting,wu2024seesr,sun2025pixel,lin2024diffbir,yu2024scaling}, pushing the boundaries of realism within the perception-distortion tradeoff \cite{blau2018perception}. 
Recent research has explored 
more complex degradations/artifacts \cite{li2022face,chen2025faithdiff,chen2024restoreagent,zhang2024degradation,wei2025perceive,ren2025hallucination}, 
higher zooms/resolutions \cite{kim2025chain,moser2024zoomed}, 
efficiency \cite{wu2024one,noroozi2024you}, 
and architectures \cite{yi2025fine,duan2025dit4sr}. 
However, naive (independent) per-frame SR
may hallucinate details inconsistent with multi-frame observations. In this work, we focus on \textit{burst} restoration, where \textit{real} details are extracted from multiple images, not purely generated. We therefore turn to video-based priors, which provide natural mechanisms for handling multiple frames. 

\noindent
\textbf{Diffusion-based SR: Videos. }
As for images, modern video restorers \cite{wang2025seedvr2,kong2025dam,xie2025star,wang2025turbovsr} have also begun to adapt large video diffusion priors \cite{yang2024cogvideox,wan2025,blattmann2023stable}
(though some  \cite{sun2025one,yang2024motion,zhou2024upscale} still utilize image priors).
Recent work has focused on several challenges, including 
motion and alignment \cite{yang2024motion,shi2022rethinking}, 
architectures \cite{zhang2024realviformer,xie2025simplegvr}, and
robustness to real-world degradations \cite{zhao2024avernet,mao2025making,chan2022investigating}. 
One critical difficulty is efficiency, as video data and diffusion-based processing are both computationally expensive, leading to the popularity of one-step models \cite{chen2025dove,sun2025one,wang2025seedvr2,zhuang2025flashvsr,liu2025ultravsr}. 
%
Unlike the image scenario, video restoration requires a sequence of frames as both the input and output, resulting in temporal consistency of generated content as an additional challenge.
In contrast, burst restorers fuse multiple frames into a single one, extracting subpixel details via subtle cues. 
Further, burst data has different motion characteristics (for both camera and scene) and generally operates in the raw image domain.
The latter difference is particularly critical, as existing priors are trained in standard RGB (sRGB) \cite{yang2024cogvideox,wan2025,blattmann2023stable}.
To our knowledge, previous work \cite{tokoro2024burst,kawai2025efficient} has considered combining non-generative burst restorers with diffusion, {but} only applies an \textit{image} diffusion model to the single-frame output, to add missing details.
In contrast, our BurstGP is fundamentally multiframe-aware, built upon a video diffusion model and finetuned on bursts.

\section{Proposed Method}
\label{sec:method}

In this section, we present BurstGP, our novel approach to incorporating generative priors into raw BISR. 
The proposed method builds upon established techniques in the field, while introducing key innovations to enhance perceptual quality while allowing a controllable trade-off with reconstruction fidelity. Our framework integrates two fundamental components: (i) a conventional raw BISR architecture from recent works~\cite{kang2024burstm, luo2022bsrt, dudhane2022burst}, and (ii) a diffusion-based sRGB video SR approach~\cite{chen2025dove}, adapted and optimized for our specific application. The first innovation (Section~\ref{sec:cond}) we introduce is a novel degradation conditioning mechanism. This component plays a critical role in enhancing the fidelity of the diffusion model, 
by explicitly informing the model about the degradation level of the input.
The second major contribution (Sec.~\ref{sec:isp_inv}) is our robust ISP inversion strategy. Since the diffusion model operates in the sRGB domain while the target output of raw BISR is defined in linear RGB space, the ISP pipeline must be inverted which is a challenging problem due to the non-invertibility of key ISP operations, particularly tone-mapping, gamma compression, and clipping. Our solution employs an optimization-based inversion process that effectively approximates the reverse transformations. It is robust to saturated and dark pixels where the inversion is ill-posed, improving reconstruction of the original scene radiance. 
This enables BurstGP to move between the lRGB and sRGB domains \textit{without} extensive training on lRGB data.

\subsection{Overall Pipeline}
Our proposed methodology is inspired by DiffBIR~\cite{lin2024diffbir}, which enhances conventional image restoration outputs by fine-tuning a pretrained generative prior. Building on this, our pipeline aims to elevate the perceptual quality of BISR results with a video diffusion model, leveraging its strong spatiotemporal prior 
as a conditional refiner rather than a primary reconstructor. A critical challenge, however, stems from the domain discrepancy: pretrained diffusion models~\cite{rombach2022high, yang2024cogvideox, wan2025} are typically trained on sRGB data, while BISR operates in linear RGB (lRGB) space. This necessitates a framework to align these distinct domains. Moreover, unlike standard video SR, where the diffusion model consumes and returns a temporal clip, BISR methods restore a single frame from the entire burst.
{One straightforward solution is to utilize a diffusion-based single-image SR model to enhance the BISR output, as in BSRD \cite{tokoro2024burst}. 
However, this does not take advantage of the constraints provided by the multi-frame input.
}
Thus, we propose to exploit the multiframe-aware nature of \textit{video} SR models. 

\begin{figure}[tb]
  \centering
  \includegraphics[width=\textwidth]{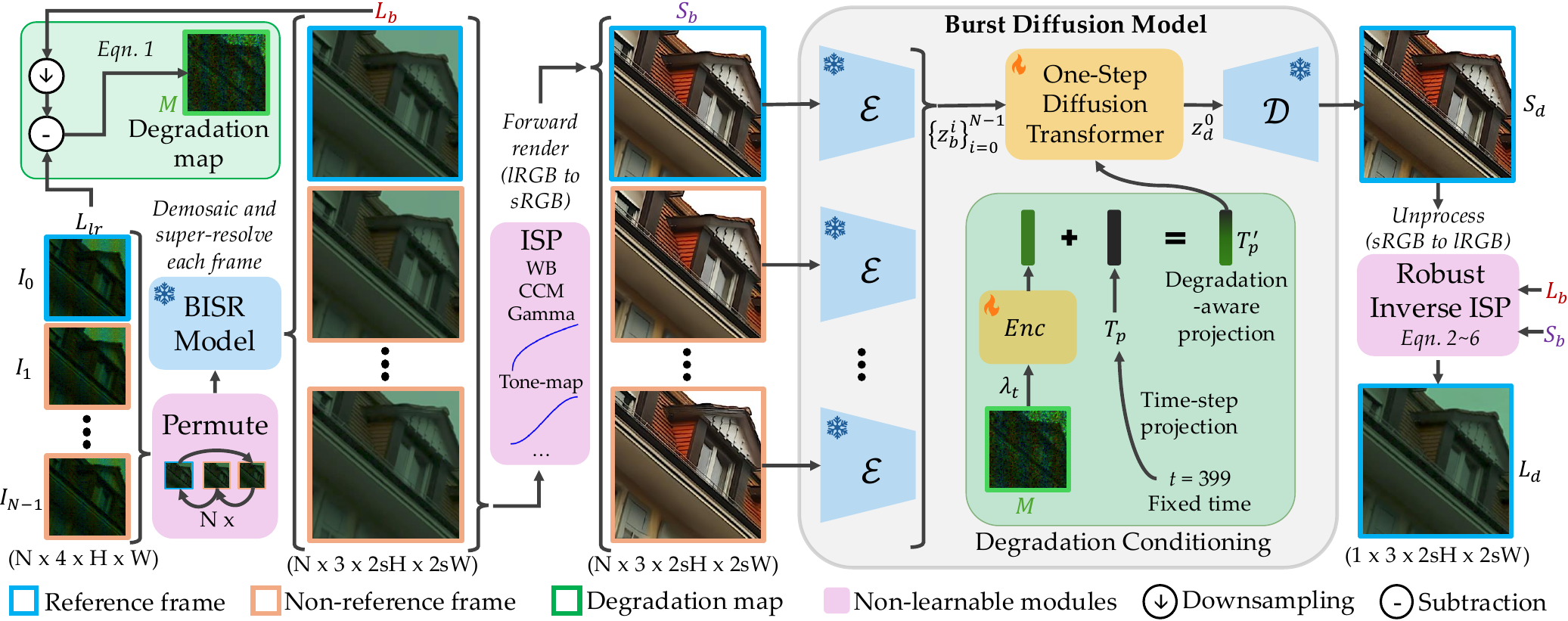}
  \caption{
  We begin with super-resolving a raw LR burst $\{I_i\}_{i=0}^{N-1}$ into a linear RGB (lRGB) SR burst using a Burst Image Super-Resolution (BISR) model.
  We super-resolve each frame separately by assuming it to be the reference frame and treating others as source frames. Let the LR and SR versions of the actual reference frame be $L_{lr}$ and $L_{b}$. We render the lRGB SR burst in sRGB color-space using an ISP module, denoted $S_b$, which is then passed to the Burst Diffusion Model. The encoder $\mathcal{E}$ processes each input frame individually to produce latents, $\{z^i_b\}^{N-1}_{i=0}$, which a one-step diffusion transformer then restores to $\{z^i_d\}^{N-1}_{i=0}$. The decoder, $\mathcal{D}$, decodes the latent, $z^0_d$, into a high-quality sRGB SR reference frame, $S_d$. To inform the diffusion model about the level of degradation in the input, we replace the original sinusoidal time projection, $T_p$, with a degradation-aware version, $T'_p$, which combines $T_p$ and an encoding of the degradation map, $M$ (see Sec.~\ref{sec:cond}). We transform the output, $S_d$, into the lRGB color-space using a Robust Inverse ISP module, which also ingests $L_b$ and $S_b$ (see Sec.~\ref{sec:isp_inv}).
  Without loss of generality, we use bursts of length $N=14$, as in other works \cite{kang2024burstm,luo2022bsrt}.
  }
  \label{fig:pipeline}
\end{figure}

As illustrated in Fig.~\ref{fig:pipeline}, our approach begins with a conventional BISR model applied to the burst sequence, $\{I_i\}_{i=0}^{N-1}$, containing $N$ frames to reconstruct a HR reference frame $L_b$ in the linear RGB space. To extend HR reconstruction to all frames in the burst, we generate circular permutations of the burst sequence and process each permuted burst independently. The resulting lRGB outputs of all frames from the BISR model are then transformed into sRGB using a standard image signal processing (ISP) pipeline~\cite{kang2024burstm, dudhane2023burstormer, brooks2019unprocessing}. We denote the super-resolved sRGB reference frame as $S_b$. Next, we leverage DOVE~\cite{chen2025dove}, a one-step video diffusion model, as our pretrained generative prior for fine-tuning,  
as it maintains a strong realism-fidelity balance, while also being efficient.
DOVE processes the sRGB frames to produce an enhanced reference frame, $S_d$. Adhering to its stage-2 training protocol, we encode frames individually to avoid temporal compression and preserve finer detail. The diffusion model is fine-tuned using Low-Rank Adaptation (LoRA)~\cite{hu2022lora}, with supervision applied exclusively to the reference frame to align with our objective. Experimental results in Sec.~\ref{sec:exp}  validate that using the multi-frame diffusion model substantially improves upon single-image diffusion methods in refining the super-resolved reference frame. 

To enhance controllability, we introduce a novel degradation conditioning mechanism (Sec.\ \ref{sec:cond}): a degradation map derived from the initial BISR restoration and LR inputs is injected into the diffusion model’s time embedding.
This controls the model's generative capabilities based on the amount of degradation, thereby improving perceptual quality while limiting fidelity loss.
Finally, since the diffusion model operates in sRGB, but our target output resides in lRGB, we introduce a robust ISP inversion strategy (Sec.\ \ref{sec:isp_inv}) to invert $S_d$ back to the linear domain, producing $L_d$. This step addresses the non-invertibility of the ISP, 
ensuring accurate reconstruction of the final linear HR image.

\subsection{Degradation Conditioning}
\label{sec:cond}

The time embedding module in DOVE follows established practices~\cite{yang2024cogvideox, peebles2023scalable}, converting diffusion timesteps into conditioning vectors via sinusoidal encoding and an MLP. These vectors are globally injected through adaptive layer normalization, modulating both normalization statistics and residual branches in attention and MLP layers. The timestep, $t$, encodes the noise level of intermediate samples, governing the signal-to-noise ratio (SNR) and controlling the strength of the generative prior. Notably, DOVE initializes its diffusion process using VAE-encoded LR features as noisy latents.
Further, DOVE’s pretraining employs a fixed timestep, $t=399$, which fails to adapt to varying degradation levels. This is suboptimal for our task, as the BISR model’s preliminary restoration significantly suppresses noise and blur, suggesting that a smaller $t$ may better preserve fidelity. Yet, timestep selection remains heuristic- and dataset-dependent.

To address this, we propose a novel degradation conditioning mechanism (Fig.~\ref{fig:pipeline}) that preserves DOVE’s pretrained patterns while dynamically adjusting the time embedding. Given the BISR-restored reference frame, $L_b$, we: (i) downsample it to the original LR resolution and mosaic it to raw format; (ii) subtract it from the LR reference frame, $L_{lr}$, to get the degradation map, $M$; and (iii) encode the degradation map via a learnable lightweight convnet and global average pooling to obtain a conditioning vector, which is injected into the timestep sinusoidal projection, $T_p$, as:
\begin{equation}
M \leftarrow L_{lr} - Mos (Down({L_{b}})),\quad T'_p \leftarrow T_p + Enc(\lambda_t M) ,\quad E \leftarrow Emb(T'_p),
\label{eq:time_embed} 
\end{equation}

\noindent
where $Emb$, $Enc$, $Mos$, and $Down$ denote the time embedder, CNN encoder, mosaicker, and downsampler, respectively. $E$ represents the updated time embedding, $M$ represents the degradation map, and $\lambda_t$ acts as a scale factor that controls the magnitude of $M$. 
{In summary, our approach is to replace the timestep projection of the original model, $T_p$, with a degradation-aware modification, $T_p'$.} 

BISR methods tend to produce over-smoothed reconstructions in challenging regions (\eg, high-noise areas), as they prioritize fidelity over perceptual quality. We quantify this using the degradation map, $M$, 
which measures the magnitude of discrepancy between the LR input and BISR reconstruction.
Injecting this signal into the timestep pathway of the time embedding via degradation conditioning not only eliminates reliance on a fixed heuristic timestep, $t$, but also enables the diffusion model to rely more strongly on its pretrained generative prior when the BISR output is unreliable. 

During fine-tuning, we supervise sRGB DOVE outputs, $S_{d}$, against HR sRGB targets, $S_{hr}$, rendered from the GT, $L_{hr}$, in linear space. We randomly set $\lambda_t = 0$ for 10$\%$ of samples, for which we replace $S_{hr}$ with $S_b$ (rendered from $L_b$), and keep $\lambda_t = 1$ and $S_{hr}$ as the GT for the remaining samples. This strategy encourages the diffusion model to adapt to different levels of ambiguity. 
Specifically, when $\lambda_t = 0$, the model is encouraged to replicate ``perfect'' BISR reconstructions, in idealized scenarios with minimal degradation. Conversely, when $\lambda_t = 1$, it prioritizes high perceptual quality. 
By tuning $\lambda_t \in [0,1]$, we explicitly move between emphasizing fidelity vs.\ perceptual quality
(see Sec.~\ref{sec:trade_off}).
By default, we set $\lambda_t = 1$.
Our ablation in
Sec.~\ref{sec:ablations} further shows that degradation conditioning improves fidelity, while maintaining perceptual quality. 

\subsection{Robust Inverse ISP}
\label{sec:isp_inv}

After obtaining the enhanced sRGB results, $S_{d}$, with DOVE,  we convert it back to lRGB space to produce, $L_{d}$, aligning with the objective of raw BISR. Let $F_{ISP}$ denote the image signal processor (ISP) function, which maps lRGB inputs to sRGB. A direct global inversion can be described via $L_d\approx F_{ISP}^{-1}(S_d)$,
\noindent
which is infeasible due to the non-invertibility of $F_{ISP}$, primarily caused by clipping operations after white-balancing, as well as the non-linear tone mapping and gamma compression curves. In particular, the tone-mapping curve ($f_t(x) = 3x^2 - 2x^3$, in our case~\cite{brooks2019unprocessing})
exhibits vanishing gradients in near-dark regions and saturated highlights, while the gamma compression curve ($f_g(x) = x^{1 / 2.2}$) shows steep gradients at near-dark intensities, rendering the inversion ill-posed and numerically unstable. Consequently, minor deviations in $S_d$, which may lie outside the ISP manifold due to the generative nature of diffusion models, can induce significant errors in $L_d$. This necessitates a more robust inversion strategy.

To circumvent the non-invertibility of $F_{ISP}$ and make use of perceptual diffusion priors, we introduce a pixel-wise residual inversion framework where each pixel is processed independently. Our method adopts a two-stage correction scheme that combines first-order residual transport with a low-rank truncated-SVD (TSVD) refinement. We compute $\Delta L$ relative to $L_b$, instead of directly estimating $L_d$. In the first stage, we leverage a Taylor approximation to compute a first-order update, formulated as:%
\begin{gather}%
S_d = F_{ISP}(L_{b} + \Delta L) \approx F_{ISP}(L_{b}) + F_{ISP}'(L_{b})\Delta L = S_{b} + F_{ISP}'(L_{b})\Delta L \\ 
\Delta L \approx \arg\!\min_{\Delta L} ||F_{ISP}'(L_{b})\Delta L - \Delta S||^2 + \beta||\Delta L||^2, \quad \Delta S \leftarrow S_d - S_b,
\label{eq:tylor} 
\end{gather}

\noindent
where $\beta=10^{-6}$ is a regularization weight 
to stabilize the solution. 
Let $J\in \mathbb{R}^{3 \times 3}$ denote the per-pixel Jacobian matrix, $F_{ISP}'(L_b)$, 
obtained by analytically differentiating the ISP pipeline and evaluating the resulting derivatives at $L_b$. Then the problem corresponds to a linear regression~\cite{tikhonov_solutions_1977} with closed-form solution
\begin{equation}
\Delta L_{fo} \approx (J^\intercal J + \beta I)^{-1}J^\intercal\Delta S,
\label{eq:close_fo} 
\end{equation}

\noindent
where $I$ represents the identity matrix. While stable for most pixels, this update may still 
introduce significant drift 
whenever
$J$ is rank-deficient (\eg, in near-dark and saturated regions). Note that all pixels are solved independently.

To prevent this drift of $\Delta L_{fo}$, in the second stage, we perform a TSVD refinement, limited to ill-conditioned pixels. Specifically, we compute the truncated decomposition of $J$ as $U_k\Sigma_k V_k^\intercal$, retaining $k$ singular values above a threshold, and estimate $\Delta L$ following~\cite{hansen1987truncated} as
\begin{equation}
\Delta L_{TSVD} \approx V_k (\Sigma_k^\intercal\Sigma_k + \beta I)^{-1}\Sigma_k^\intercal U_k^\intercal \Delta S.
\label{eq:close_tsvd} 
\end{equation}

\noindent
This TSVD refinement strictly suppresses updates in the nullspace of $J$, where the ISP provides no physical observability (\ie, when the inversion is ill-conditioned). The final residual update is given by 
\begin{equation}
\Delta L = m\Delta L_{fo} + (1 - m)\Delta L_{TSVD}, \quad L_d = L_b + \lambda_r\Delta L,
\label{eq:delta_l} 
\end{equation}

\noindent
where $m = 1$ for well-conditioned pixels and $m=0$ for the rest. The scale factor $\lambda_r\in[0,1]$  controls the magnitude of the residual update, which we set to $\lambda_r = 1$ by default. By tuning the scale factor $\lambda_r$ in the final lRGB outputs, we can also adjust the trade-off between fidelity and perceptual quality, providing an additional advantage of residual inversion over direct inversion (see Sec.~\ref{sec:trade_off}).
More detailed mathematical derivations are provided in the supplement.

\section{Experiments}
\label{sec:exp}

\subsection{Implementation Details}
\label{sec:details}

\noindent
\textbf{Synthetic Data}. Following previous work~\cite{bhat2021deep}, we generate the SyntheticBurst dataset with an inverse camera pipeline~\cite{brooks2019unprocessing} 
and downsampling the Zurich RAW to RGB dataset~\cite{ignatov2020replacing}, which has 46,839 training and 1,204 testing cropped patches. Each burst contains $N=14$ LR raw inputs and one HR lRGB ground-truth.

\noindent
\textbf{Real Data}. The BurstSR dataset~\cite{bhat2021deep} consists of LR-HR pairs captured with different devices (again with $N=14)$. 
We train the model on 5,405 training patches and evaluate on 882 testing patches. 
RealBSR-RAW~\cite{wei2023towards} includes 20,842 training and 2,377 testing bursts (14-frame $160{\times}160$ LR crops for $4\times$ SR), with LR and HR captured via optical zoom on the same device.

\noindent
\textbf{Training Settings}. 
We initialize DOVE and BISR models using official checkpoints (DOVE’s stage-2 weights) and fine-tune on eight NVIDIA A100-80G GPUs for 5,000 iterations with batch size one per GPU, a learning rate of $5\times 10^{-5}$ and the AdamW optimizer~\cite{loshchilov2017fixing}. 
For our losses during dataset-specific fine-tuning, on the SyntheticBurst dataset, we use the original loss from DOVE, since the LR-HR pairs are perfectly aligned.
Note that to address interpolation artifacts in SyntheticBurst’s non-reference frames, we retrain a dedicated BISR model with random burst permutation during synthetic training, enhancing robustness to domain gaps between reference and non-reference frames.
On the real-world BurstSR dataset, following prior work~\cite{kang2024burstm}, we employ the aligned $L_1$ loss. 
Finally, on RealBSR-RAW, we follow FBANet~\cite{wei2023towards} and combine an $L_1$ term with the CoBi loss~\cite{zhang2019zoom}. 
Additional details are provided in the supplement. 

\noindent
\textbf{Evaluation}. 
To evaluate BurstGP, we compare against six state-of-the-art BISR approaches on SyntheticBurst and BurstSR, including EBSR~\cite{luo2021ebsr}, DBSR~\cite{bhat2021deep}, MFIR~\cite{bhat2021deep2}, BIPNet~\cite{dudhane2022burst}, BSRT~\cite{luo2022bsrt} (both small and large), and BurstM~\cite{kang2024burstm}. We evaluate via inference of officially released checkpoints. For quantitative evaluation, we adopt PSNR and SSIM in both lRGB and sRGB spaces (the latter rendered from lRGB outputs with the given ISP), alongside perceptual metrics LPIPS~\cite{zhang2018unreasonable}, TOPIQ~\cite{chen2024topiq}, and MUSIQ~\cite{ke2021musiq} in sRGB space. 
On RealBSR-RAW, we compare to BurstM~\cite{kang2024burstm} and FBANet~\cite{wei2023towards} (trained ourselves, as public checkpoints are unavailable).
{On real data, we also include shift-tolerant LPIPS (ST-LPIPS)~\cite{ghildyal2022shift} to address potential misalignments with respect to the ground-truth.} 
This suite of metrics
ensures comprehensive analysis of fidelity and perceptual quality. 
Finally, we remark that we do not compare to BSRD~\cite{tokoro2024burst}, as they operate on lower resolutions ($256^2$ vs.\ $384^2$ on SyntheticBurst) and smaller bursts (eight vs.\ fourteen frames), compared to existing state-of-the-art BISR models.

\subsection{Empirical Results}

\noindent
\textbf{Quantitative Evaluation}. 
As shown in Tab.~\ref{tab:syn_quan}, our BurstGP significantly improves perceptual quality on the SyntheticBurst dataset compared to BISR baselines, albeit with a fidelity trade-off that can be effectively compensated through tuning $\lambda_t$ and $\lambda_r$ (Sec.~\ref{sec:trade_off}). On the BurstSR dataset (Tab.~\ref{tab:bsr_quan}), our method outperforms baselines in \textit{both} fidelity and perceptual quality. We attribute this to the saturation of baseline models in linear space (PSNR ${\sim}$50 dB), where minor color mismatches and distortion become negligible. However, when converted to sRGB, these discrepancies are amplified, enabling our approach to further refine the output quality. Our results on RealBSR-RAW~\cite{wei2023towards} are shown in Tab.~\ref{tab:realbsr_quan}, showing that perceptual quality is significantly boosted by BurstGP (TOPIQ ${+}25$\% and MUSIQ ${+}9$\%), as expected.
However, we do not incur a significant loss to fidelity, 
as we are similar to FBANET
in terms of low-level metrics (${<}{\pm}2$\% change in PSNR and SSIM), while substantially improving LPIPS (${-}14$\%) and ST-LPIPS (${-}17$\%). 
In other words, we can improve no-reference quality \textit{and} perceptual fidelity, with only marginal cost to pixel-level accuracy.    

\begin{table}[t]
	\caption{Quantitative evaluations on the SyntheticBurst dataset.}
	\footnotesize
	\centering
	\begin{adjustbox}{width=0.83\linewidth}
\begin{tabular}{lccccccc}
\toprule
Method         & PSNR$~\uparrow$ & PSNR-L$~\uparrow$ & SSIM$~\uparrow$ & SSIM-L$~\uparrow$ & LPIPS$~\downarrow$ & TOPIQ$~\uparrow$ & MUSIQ$~\uparrow$\\ 
\midrule
Bicubic & 19.90 & 30.15 & 0.6437 & 0.8035 & 0.6449 & 0.2563 & 16.73  \\
DBSR~\cite{bhat2021deep} & 31.66 & 38.88 & 0.8851 & 0.9377 & 0.3077 & 0.6991 & 36.31 \\
MFIR~\cite{bhat2021deep2} & 34.25 & 41.26 & 0.9245 & 0.9602 & 0.2569 & 0.8094 & 42.52 \\
EBSR~\cite{luo2021ebsr} & \underline{35.75} & \underline{42.65} & 0.9387 & 0.9684 & 0.2253 & 0.8471 & 44.78  \\

\rowcolor{purple!10}BIPNet~\cite{dudhane2022burst} & 35.42 & 42.41 & \underline{0.9392} & 0.9683 & 0.2270 & 0.8443 & 43.72  \\
\rowcolor{purple!10} \textbf{+BurstGP(ours)} & 33.43 & 40.12 & 0.9157 & 0.9574 & 0.1712 & 0.8749 & 45.52 \\

\rowcolor{cyan!15}BurstM~\cite{kang2024burstm} & 35.63 & 42.47 & 0.9383 & 0.9680 & 0.2293 & 0.8437 & 44.66  \\
\rowcolor{cyan!15} \textbf{+BurstGP(ours)} & 34.02 & 40.71 & 0.9165 & 0.9576 & \underline{0.1618} & 0.8836  & 45.57  \\

\rowcolor{blue!15}BSRT-S~\cite{luo2022bsrt} & 35.51 & 42.50 & 0.9391 & \underline{0.9686} & 0.2275 & 0.8473 & 44.48  \\
\rowcolor{blue!15} \textbf{+BurstGP(ours)} & 33.95 & 40.69 & 0.9158 & 0.9574 & 0.1634 & \underline{0.8853} & \underline{45.70}  \\

\rowcolor{violet!15}BSRT-L~\cite{luo2022bsrt} & \textbf{36.55} & \textbf{43.47} & \textbf{0.9462} & \textbf{0.9731} & 0.2114 & 0.8656 & 45.69  \\
\rowcolor{violet!15} \textbf{+BurstGP(ours)} & 34.29 & 41.08 & 0.9194 & 0.9602 & \textbf{0.1543} & \textbf{0.8913} & \textbf{46.72} \\

\bottomrule
\end{tabular}
	\end{adjustbox}
	\label{tab:syn_quan}
\end{table}
\begin{table}[t]
	\caption{
        Quantitative evaluations on the BurstSR (real) dataset.
        Following prior work~\cite{luo2022bsrt}, all full-reference metrics are computed post-alignment, except ST-LPIPS.
    }
	\footnotesize
	\centering
	\begin{adjustbox}{width=\linewidth}
\begin{tabular}{lcccccccc}
\toprule
Method         & PSNR$~\uparrow$ & PSNR-L$~\uparrow$ & SSIM$~\uparrow$ & SSIM-L$~\uparrow$ & LPIPS$~\downarrow$ & ST-LPIPS$~\downarrow$ & TOPIQ$~\uparrow$ & MUSIQ$~\uparrow$\\ 
\midrule
Bicubic & 25.21 & 42.73 & 0.7640 & 0.9492 & 0.5018 & 0.4573 & 0.3614 & 18.59 \\
DBSR~\cite{bhat2021deep} & 29.88 & 47.77 & 0.8885 & 0.9809 & 0.3507 & 0.2359 & 0.6284 & 38.75 \\
MFIR~\cite{bhat2021deep2} & 30.37 & 48.39 & 0.9015 & 0.9828 & 0.3280 & 0.2100 & 0.6784 & 42.70 \\
EBSR~\cite{luo2021ebsr} & 30.40 & 48.30 & 0.8996 & 0.9826 & 0.3158 & 0.2059 & 0.6777 & 43.56 \\

\rowcolor{purple!10}BIPNet~\cite{dudhane2022burst} & 29.91 & 47.94 & 0.8924 & 0.9815 & 0.3337 & 0.2289 & 0.6380 & 38.42 \\
\rowcolor{purple!10}\textbf{+BurstGP(ours)} & 30.61 & 48.27 & 0.9010 & 0.9829 & 0.3059 & 0.2101 & 0.6980 & 43.65  \\
\rowcolor{cyan!15}BurstM~\cite{kang2024burstm} & 30.31 & 48.24 & 0.8989 & 0.9824 & 0.3231 & 0.2076 & 0.6725 & 43.57 \\
\rowcolor{cyan!15}\textbf{+BurstGP(ours)} & 30.88 & 48.42 & \underline{0.9052} & \underline{0.9837} & 0.3000 & 0.1962 & 0.7105 & 46.48  \\
\rowcolor{blue!15}BSRT-S~\cite{luo2022bsrt} & 30.53 & 48.49 & 0.9034 & 0.9832 & 0.3190 & 0.2043 & 0.6873 & 43.16 \\
\rowcolor{blue!15}\textbf{+BurstGP(ours)} & \underline{30.98} & \underline{48.62} & \textbf{0.9080} & \textbf{0.9839} & \underline{0.2976} & \underline{0.1903} & \underline{0.7238} & \underline{47.28} \\
\rowcolor{violet!15}BSRT-L~\cite{luo2022bsrt} & 30.59 & 48.59 & 0.9050 & 0.9835 & 0.3111 & 0.1981 & 0.6976 & 44.44 \\
\rowcolor{violet!15}\textbf{+BurstGP(ours)} & \textbf{30.99} & \textbf{48.64} & \textbf{0.9080} & \textbf{0.9839} & \textbf{0.2960} & \textbf{0.1881} & \textbf{0.7267} & \textbf{47.89} \\

\bottomrule
\end{tabular}
	\end{adjustbox}
	\label{tab:bsr_quan}
\end{table}
\begin{table}[t]
	\caption{Evaluation on the RealBSR-RAW dataset \cite{wei2023towards}. Following the dataset authors, we do not use additional post-processing in evaluation. 
    }
	\footnotesize
	\centering
\resizebox{\textwidth}{!}{
\begin{tabular}{lcccccccc}
\toprule
Method         & PSNR$~\uparrow$ & PSNR-L$~\uparrow$ & SSIM$~\uparrow$ & SSIM-L$~\uparrow$ & LPIPS$~\downarrow$ & ST-LPIPS$~\downarrow$ & TOPIQ$~\uparrow$ & MUSIQ$~\uparrow$ \\ 
\midrule
Bicubic &     
22.19 & 28.93 & 0.6989 & 0.7709 & 0.4258 & 0.3022 & 0.2391 & 31.51 \\
BurstM \cite{kang2024burstm} &     
22.38 & \textbf{31.28} & \textbf{0.7387} & \textbf{0.8611} & 0.2944 & 0.2839 & 0.2516 & 41.31 \\ 
\rowcolor{blue!15} FBANet \cite{wei2023towards} &     
\textbf{22.67} & 31.21 & \underline{0.7085} & \underline{0.8484} & \underline{0.1760} & \underline{0.1905} & \underline{0.3076} & \underline{46.82} \\ 
\rowcolor{blue!15} \textbf{+BurstGP(ours)} & 
\underline{22.53} & \underline{31.24} & 0.6963 & 0.8462 & \textbf{0.1504} & \textbf{0.1582} & \textbf{0.3883} & \textbf{51.24} \\
\bottomrule
\end{tabular}
}
	\label{tab:realbsr_quan}
\end{table}

\begin{table}[t]
	\centering
    \caption{Controlling the perception-distortion trade-off.}
	\begin{adjustbox}{width=\linewidth}
    \begin{subtable}[t]{0.45\textwidth}
        \centering
        \begin{adjustbox}{width=\linewidth}
        \begin{tabular}{cccccc}
        \toprule
              & PSNR$~\uparrow$    & PSNR-L$~\uparrow$ & TOPIQ$~\uparrow$ & MUSIQ$~\uparrow$ \\ 
        \midrule
        BISR        &  \textbf{36.55}   &    \textbf{43.47}  &  0.8656    &   45.69          \\
        $\lambda_t=0$        &   \underline{35.02}  &    \underline{41.81}    &   0.8607   &   45.76         \\
        $\lambda_t=0.125$       &   34.60   &   41.37       &  0.8851     &    46.41        \\
        $\lambda_t=0.25$      &    34.39   &  41.16      &  0.8906    &    46.63        \\
        $\lambda_t=0.5$      &   34.30    &   41.09     &  \underline{0.8912}    &   \underline{46.69}         \\
        $\lambda_t=1.0$  & 34.29   &   41.08       &  \textbf{0.8913}   &   \textbf{46.72}       \\ 
        \bottomrule
        \end{tabular}
	   \end{adjustbox}
       \caption{Evaluations of $\lambda_t$ with $\lambda_r=1.0$}
        \label{tab:lambda_t}
    \end{subtable}
    \hfill 
    \begin{subtable}[t]{0.486\textwidth}
        \centering
        \begin{adjustbox}{width=\linewidth}
        \begin{tabular}{cccccc}
        \toprule
            & PSNR$~\uparrow$      & PSNR-L$~\uparrow$ & TOPIQ$~\uparrow$ & MUSIQ$~\uparrow$ \\ 
        \midrule
        BISR ($\lambda_r=0$)     &   \textbf{36.55}    &    \textbf{43.47}  &  0.8656    &  45.69         \\
        $\lambda_r=0.2$   &  \underline{36.53}  &  \underline{43.37}    &    0.8730     &     45.97     \\
        $\lambda_r=0.4$  &  36.23  &    43.00     &   0.8809    &      46.21        \\
        $\lambda_r=0.6$   &  35.71  &   42.44     &   0.8874   &    46.41          \\
        $\lambda_r=0.8$   & 35.04  &   41.78   &  \underline{0.8911}    &      \underline{46.58}      \\
        $\lambda_r=1.0$   & 34.29  &   41.08       &  \textbf{0.8913}   &   \textbf{46.72}            \\ 
        \bottomrule
        \end{tabular}
	\end{adjustbox}
        \caption{Evaluations of $\lambda_r$ with $\lambda_t=1.0$}
        \label{tab:lambda_r}
    \end{subtable}
	\end{adjustbox}
	\label{tab:lambda}
\end{table}

\begin{figure}[tb]
  \centering
  \begin{minipage}{\linewidth}
    \includegraphics[width=\linewidth]{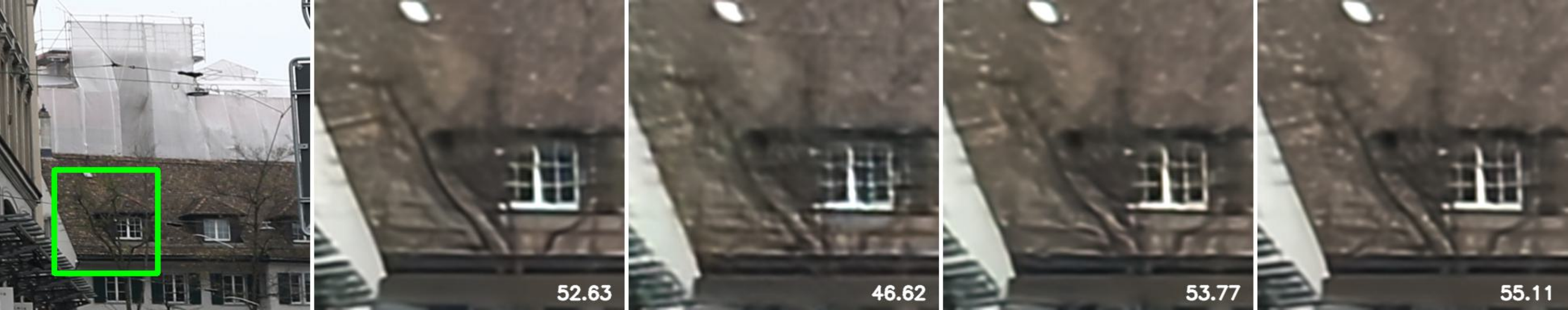}
    \vfill \vspace*{-0.12cm}
    \text{\scriptsize ~~~~~~~~~~GT~~~~~~~~~~~~~~~~~~~~BIPNet~~~~~~~~~~~~~~~~BurstM~~~~~~~~~~~~~~~BSRT-S~~~~~~~~~~~~~~~~BSRT-L}
    \vfill \vspace*{0.12cm}
  \end{minipage}
  \vfill
  \begin{minipage}{\linewidth}
    \includegraphics[width=\linewidth]{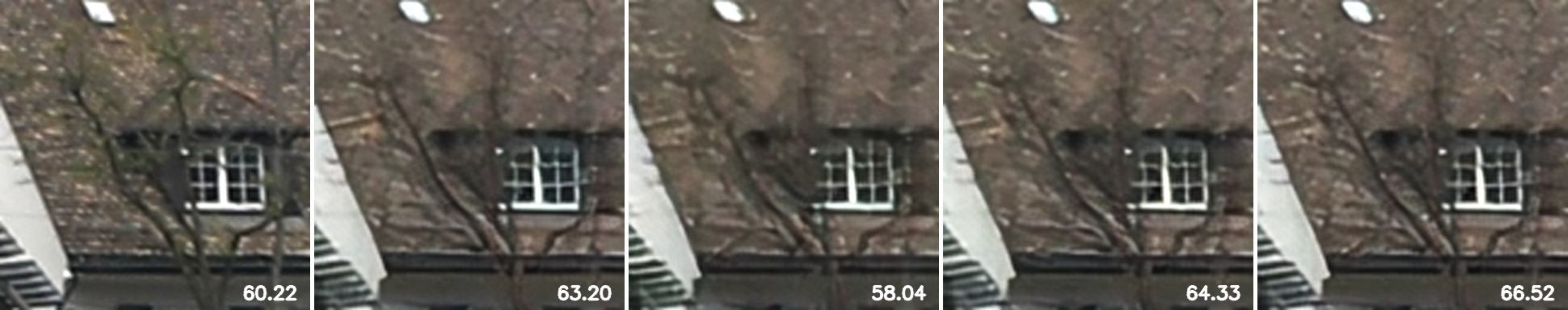}
    \vfill \vspace*{-0.12cm}
    \text{\scriptsize \hspace{2.45cm} BIPNet+BurstGP~~BurstM+BurstGP~~BSRT-S+BurstGP~~BSRT-L+BurstGP}
    \vfill \vspace*{0.12cm}
  \end{minipage}
  \vfill
  \begin{minipage}{\linewidth}
    \includegraphics[width=\linewidth]{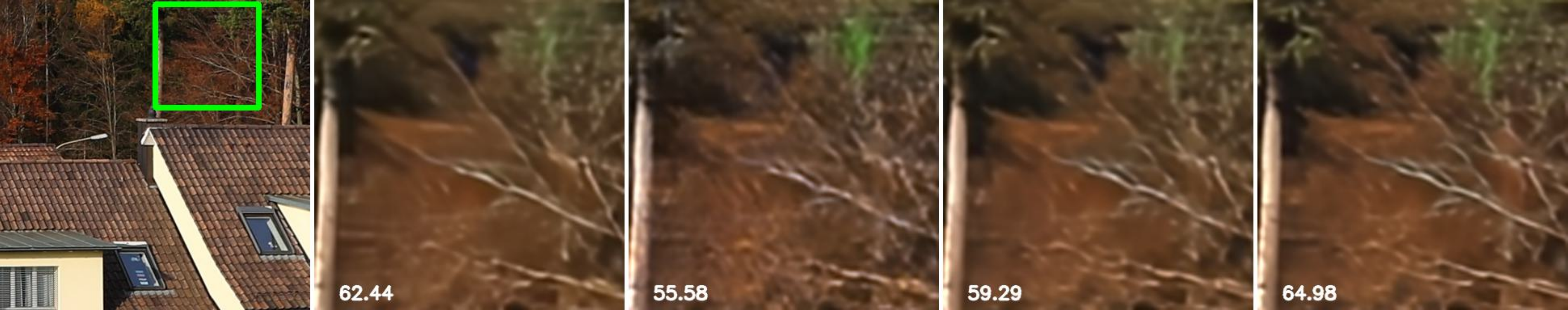}
    \vfill \vspace*{-0.12cm}
    \text{\scriptsize ~~~~~~~~~~GT~~~~~~~~~~~~~~~~~~~~BIPNet~~~~~~~~~~~~~~~~BurstM~~~~~~~~~~~~~~~BSRT-S~~~~~~~~~~~~~~~~BSRT-L}
    \vfill \vspace*{0.12cm}
  \end{minipage}
  \vfill
  \begin{minipage}{\linewidth}
    \includegraphics[width=\linewidth]{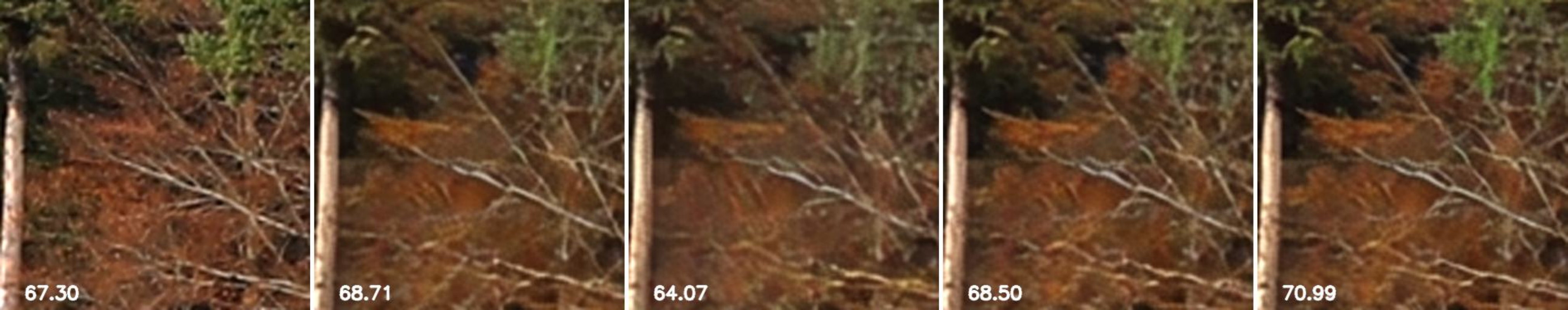}
    \vfill \vspace*{-0.12cm}
    \text{\scriptsize \hspace{2.45cm} BIPNet+BurstGP~~BurstM+BurstGP~~BSRT-S+BurstGP~~BSRT-L+BurstGP}
    \vfill \vspace*{0.12cm}
  \end{minipage}
  \caption{Visualizations on SyntheticBurst. Patches are shown at increased brightness for better visualization. We report full-image MUSIQ (bottom-left corner). 
  BurstGP consistently improves textural detail, but avoids hallucinating unrealistic content.
  }
  \label{fig:syn_qual}
\end{figure}
\begin{figure}[tb]
  \centering
  \begin{minipage}{\linewidth}
    \includegraphics[width=\linewidth]{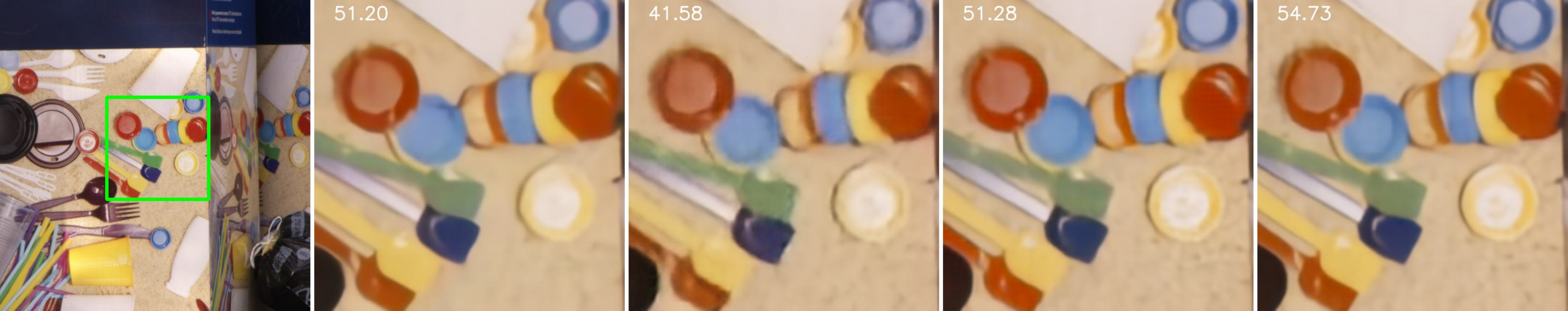}
    \vfill \vspace*{-0.12cm}
    \text{\scriptsize ~~~~~~~~~~GT~~~~~~~~~~~~~~~~~~~~BIPNet~~~~~~~~~~~~~~~~BurstM~~~~~~~~~~~~~~~BSRT-S~~~~~~~~~~~~~~~~BSRT-L}
    \vfill \vspace*{0.12cm}
  \end{minipage}
  \vfill
  \begin{minipage}{\linewidth}
    \includegraphics[width=\linewidth]{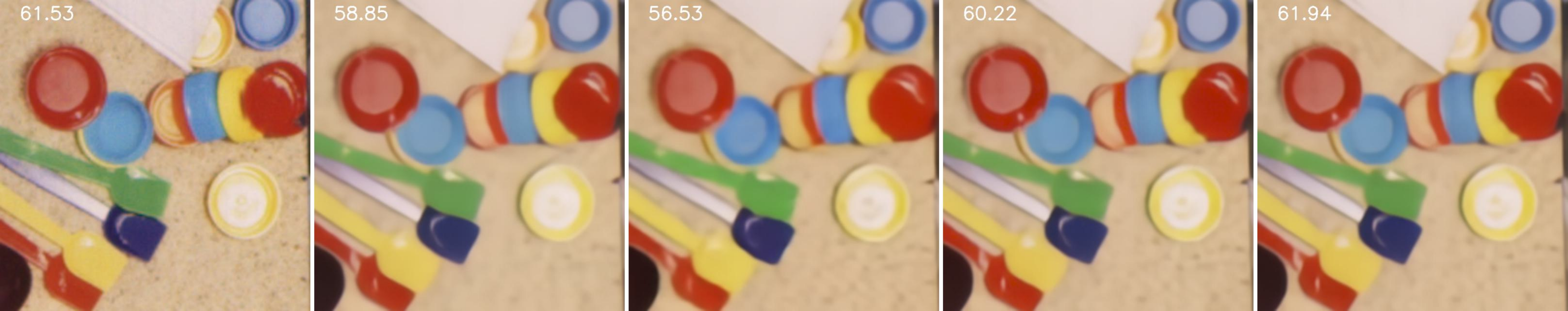}
    \vfill \vspace*{-0.12cm}
    \text{\scriptsize \hspace{2.45cm} BIPNet+BurstGP~~BurstM+BurstGP~~BSRT-S+BurstGP~~BSRT-L+BurstGP}
    \vfill \vspace*{0.12cm}
  \end{minipage}
  \vfill
  \begin{minipage}{\linewidth}
    \includegraphics[width=\linewidth]{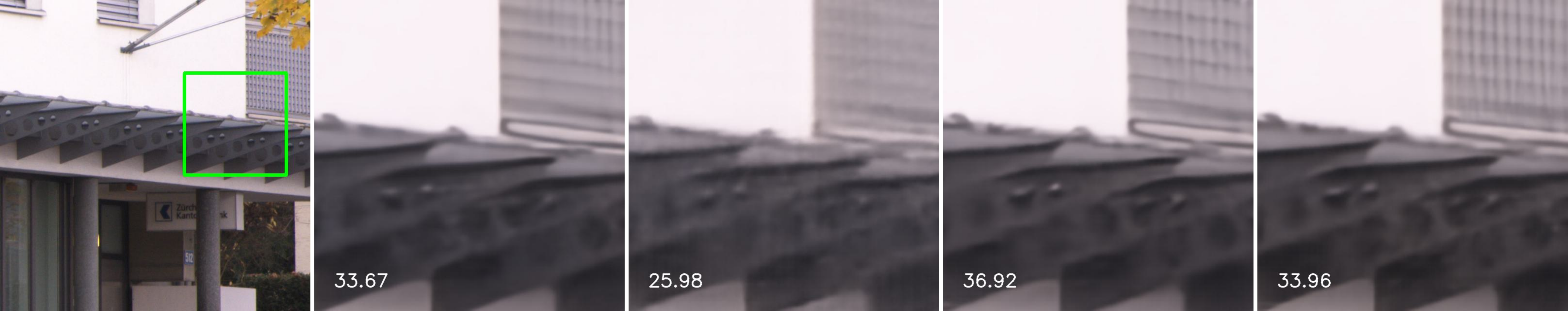}
    \vfill \vspace*{-0.12cm}
    \text{\scriptsize ~~~~~~~~~~GT~~~~~~~~~~~~~~~~~~~~BIPNet~~~~~~~~~~~~~~~~BurstM~~~~~~~~~~~~~~~BSRT-S~~~~~~~~~~~~~~~~BSRT-L}
    \vfill \vspace*{0.12cm}
  \end{minipage}
  \vfill
  \begin{minipage}{\linewidth}
    \includegraphics[width=\linewidth]{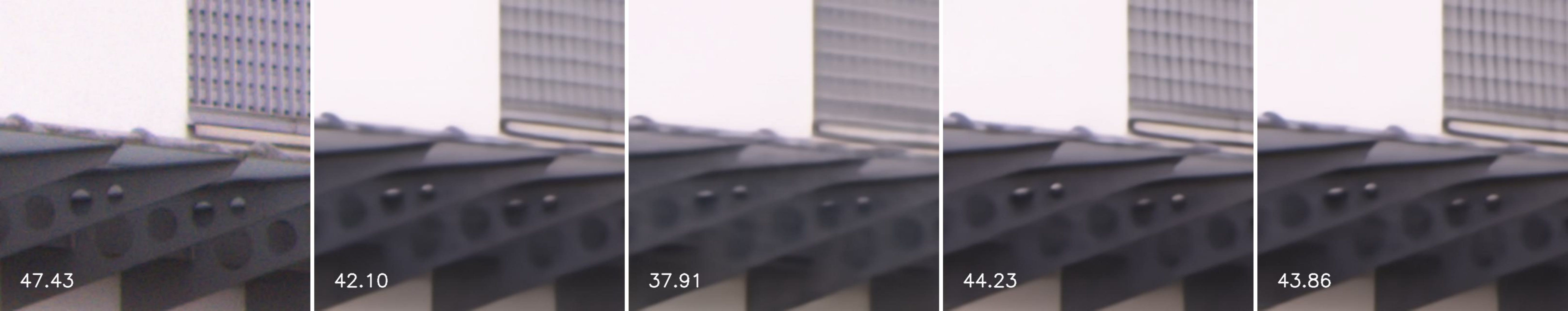}
    \vfill \vspace*{-0.12cm}
    \text{\scriptsize \hspace{2.45cm} BIPNet+BurstGP~~BurstM+BurstGP~~BSRT-S+BurstGP~~BSRT-L+BurstGP}
    \vfill \vspace*{0.12cm}
  \end{minipage}
  \caption{Qualitative results on the BurstSR dataset. We report MUSIQ for the entire image. 
  We see that BurstGP is able to reduce blur in the BISR outputs and improve colour fidelity (upper insets), improve noisy or distorted content
  (lower insets, black structure), and sometimes infer missing details (lower insets, grey grid).
  }
  \label{fig:burstsr_qual}
\end{figure}

\noindent
\textbf{Qualitative Visualizations}. 
We show qualitative results for $4\times$ SR on the SyntheticBurst and BurstSR datasets in Fig. \ref{fig:syn_qual} and \ref{fig:burstsr_qual}, respectively. For comprehensive assessment, we report MUSIQ scores for both the full SR images and GT, with zoomed-in regions to highlight fine details. Our diffusion-based BurstGP model consistently achieves higher MUSIQ scores compared to various BISR baselines, demonstrating superior perceptual quality, particularly in complex texture regions (\eg, branches, windows). Notably, BurstGP occasionally surpasses the GT in MUSIQ, likely due to the presence of noise in certain GT regions, which our method can suppress while preserving structural integrity. 

\noindent
\textbf{Timing}.
Our approach provides substantially improved visual results. However, the use of diffusion adds significant run-time overhead.
With BSRT-L as the base model (0.91s alone), we find that the
(i) initial BISR,
(ii) DOVE-based burst diffusion,
and
(iii) ISP inversion
require 1.14s, 6.59s, and 0.50s, respectively.
The majority of the additional run-time is due to the diffusion process; hence, as the efficiency of such models improves, our approach will benefit as well.

\subsection{Additional Analyses}
\label{sec:trade_off} 

\noindent
\textbf{Controlling the Perception-Distortion Trade-off.}
{In image restoration, fidelity and realism are known to be inherently competing \cite{blau2018perception}.}
To investigate this trade-off, 
in Tab.~\ref{tab:lambda},
we evaluate how quality and fidelity vary as a function of our two scale factors, $\lambda_t$ and $\lambda_r$, 
on the SyntheticBurst dataset.
For reference, we also include the results of a non-generative BISR model (BSRT-Large, our default base BISR method) in the top row.
{Recall that $\lambda_t$ implicitly signals the degradation level of the input (Sec.~\ref{sec:cond}), while $\lambda_r$ is the lRGB-space blending weight of the generative correction into the BISR output (Sec.~\ref{sec:isp_inv}).}
Traversal results are shown in Tab.~\ref{tab:lambda_t}. 
We observe that when $\lambda_t$ is smaller, the model is more conservative (higher fidelity), closer to the original BISR outputs. The impact of the generative diffusion prior becomes more noticeable as $\lambda_t$ increases, but the metrics do not change linearly, due to the non-linear nature of the degradation conditioning.
On the other hand, as shown in Tab.~\ref{tab:lambda_r}, the behaviour of $\lambda_r$ is more intuitive, showing a linear interpolation between the BISR and diffusion outputs.
Additional visualizations are provided in the supplement.




\begin{table}[t]
	\caption{Quantitative evaluation of ISP inversion.}
	\footnotesize
	\centering
	\begin{adjustbox}{width=0.8\linewidth}
\begin{tabular}{cccccccc}
\toprule
Method         & PSNR$~\uparrow$ & PSNR-L$~\uparrow$ & SSIM$~\uparrow$ & SSIM-L$~\uparrow$ & LPIPS$~\downarrow$ & TOPIQ$~\uparrow$ & MUSIQ$~\uparrow$ \\ 
\midrule
Inverse camera~\cite{brooks2019unprocessing}       &   34.20       &   39.11   &     0.9181        &   0.9460   &      0.1561       &   0.8912    & 46.70  \\ 
\rowcolor{blue!15} Ours    &    \textbf{34.29} & \textbf{41.08} & \textbf{0.9194} & \textbf{0.9602} & \textbf{0.1543} & \textbf{0.8913} & \textbf{46.72}\\
\bottomrule
\end{tabular}
	\end{adjustbox}
	\label{tab:isp}
\end{table}

\noindent
\textbf{Evaluating ISP Inversion}.
To evaluate our proposed robust ISP inversion,
we compare against the inverse camera pipeline~\cite{brooks2019unprocessing}, a widely used method for sRGB-to-lRGB conversion, which directly inverts the sRGB outputs, $S_d$, into linear space by sequentially applying inverse operations (tone mapping, gamma compression, color correction, and white balancing), assuming global bijectivity of the ISP, $F_{ISP}$. However, due to quantization, clipping, and vanishing gradients in saturated/low-signal regions (where tone derivatives approach zero), this assumption fails, leading to reconstruction error.
As quantified in Tab.~\ref{tab:isp}, \textit{our method achieves significantly higher fidelity in linear space.} Interestingly, re-rendering 
via the forward ISP exhibits a narrower performance gap in sRGB than in lRGB. This does not contradict the instability of direct inversion, but instead reflects the compressive and partially non-injective nature of the ISP. 
As a consequence, large perturbations can be attenuated, resulting in an insensitivity of sRGB to certain variations in lRGB. 
Thus, evaluating physical consistency via \textit{linear}-domain fidelity is critical, rather than relying on sRGB metrics alone.


\begin{table}[t]
	\footnotesize
	\centering
	\caption{Ablation results.}
	\begin{adjustbox}{width=\linewidth}
	\begin{tabular}{ccccccccc}
\toprule
\multirow{2}{*}{Method}        & \multicolumn{4}{c}{SyntheticBurst}                                         & \multicolumn{4}{c}{BurstSR}                                                \\ \cmidrule[0.25pt](lr){2-5} \cmidrule[0.25pt](lr){6-9} 
                              & PSNR$~\uparrow$  & PSNR-L$~\uparrow$ & TOPIQ$~\uparrow$ & MUSIQ$~\uparrow$ & PSNR$~\uparrow$ & PSNR-L$~\uparrow$ & TOPIQ$~\uparrow$ & MUSIQ$~\uparrow$ \\ 
\midrule
Ours w/ single-frame diffusion & 34.11 & 40.91 & 0.8873 & 46.42 & \underline{30.89} & \underline{48.57} & 0.7169 & 47.13\\ 
Ours w/o degradation conditioning & \underline{34.18} & \underline{41.01} & \underline{0.8906} & 46.64 & 30.81 & 48.43 & \underline{0.7250} & 47.87 \\ 
Ours w/o ISP  & 30.41 & 40.13 & 0.8168 & 44.93 & 30.24 & 47.79 & 0.6589 & 39.46 \\ 
Ours w/o fine-tuning   & 28.89  & 36.40 & 0.7500 & \textbf{60.42} & 27.28 & 44.14 & 0.6323 & \textbf{64.46}\\ 
Ours w/o permutation training  & 31.03 & 38.00 & 0.8402 & 46.47 & - & - & - & -\\ 
\rowcolor{blue!15} Ours & \textbf{34.29} &   \textbf{41.08}   &  \textbf{0.8913}   &   \underline{46.72}  & \textbf{30.99} & \textbf{48.64} & \textbf{0.7267} & \underline{47.89} \\ 
\bottomrule
\end{tabular}
	\end{adjustbox}
	\label{tab:ablation}
\end{table}

\noindent
\textbf{Ablation Studies}
\label{sec:ablations}
In this section, we perform ablations to analyze the contribution of each component of our method (see Tab.~\ref{tab:ablation}). First, we evaluate the impact of adopting a diffusion-based video super-resolution model as a conventional BISR enhancer. Specifically, we fine-tune the same diffusion model using only a \textit{single} frame as input, which effectively reduces it to a single-image enhancement model. Compared with the first row of Tab.~\ref{tab:ablation}, the multi-frame formulation provides clear performance improvements, demonstrating that cross-frame constraints introduce complementary information beyond single-frame reconstruction.
Second, we investigate the effect of degradation conditioning by fine-tuning the diffusion model without this component. 
We find that the proposed degradation conditioning not only improves image quality (second row),
but also enables a controllable perception–distortion trade-off (as shown in Tab.~\ref{tab:lambda_t} and Fig.~\ref{fig:teaser}). 
Next, we directly fine-tune the diffusion model in lRGB space without ISP conversion (third row). 
The performance drops significantly, 
which is unsurprising, as the diffusion backbone is pretrained in sRGB and adapting the model to a new image domain would require substantially larger training datasets. 
We further evaluate the pretrained diffusion model without any fine-tuning (fourth row). Although this variant achieves the highest MUSIQ score, it produces severe artifacts (as shown in supplement) and exhibits significantly worse fidelity on both datasets.
Finally, as discussed in Sec.~\ref{sec:details}, we retrain the BISR model with frame permutation to reduce artifacts on non-reference frames when evaluated on the SyntheticBurst dataset. The results in the fifth row highlight the necessity of permutation-based training. Additional  results are provided in the supplement.

\section{Conclusion}
We introduced BurstGP, a novel diffusion-based framework for raw burst image super-resolution that combines a non-generative BISR model with a generative video super resolution approach. Our key innovations include: (i) a degradation-aware conditioning mechanism that dynamically adjusts the diffusion timestep embedding by injecting a learned degradation map, enabling control over detail synthesis; and (ii) a robust ISP inversion strategy that overcomes the instability of direct inversion, while significantly improving reconstruction fidelity. Extensive experiments demonstrated that BurstGP achieves state-of-the-art performance in both quantitative metrics and perceptual quality.  Overall, our results highlight the potential of our approach to bridge the gap between generative and non-generative super-resolution paradigms.



%
%

\clearpage

\title{BurstGP: Enhancing Raw Burst Image Super Resolution with Generative Priors \\ --Supplementary Document--} 

\titlerunning{BurstGP}

\author{Dong Huo\inst{1*} \and
Tristan Aumentado-Armstrong\inst{1*} \and
Samrudhdhi B. Rangrej\inst{1*} \and
Maitreya Suin\inst{1} \and
Angela Ning Ye\inst{1} \and
Zhiming Hu\inst{1} \and
Amanpreet Walia\inst{1} \and
Amirhossein Kazerouni\inst{1,2,3,4} \and
Konstantinos G. Derpanis\inst{1,2,3,5}
Iqbal Mohomed\inst{1} \and
Alex Levinshtein\inst{1}
}

\authorrunning{D.~Huo et al.}

\institute{\mbox{AI Center - Toronto, Samsung Electronics \and 
University of Toronto} \and 
\mbox{Vector Institute \and 
University Health Network \and 
York University} \\
\textsuperscript{*} Equal contribution}

\maketitle

\setcounter{equation}{6}
\setcounter{table}{6}
\setcounter{figure}{4}
\setcounter{section}{5}

\appendix

\section{Overview}

This supplement provides additional details of our work. In Sec.~\ref{supp:sec:invisp}, we provide a detailed derivation, error analysis and a complete algorithm of the robust inverse ISP.
We remark on linear vs sRGB fine-tuning in Sec.~\ref{supp:sec:remark} (in particular, on the difficulties of training through an inverse camera pipeline).
Sec.~\ref{supp:sec:newresults} contains additional results, including qualitative visualizations and other analyses.
Finally, in Sec.~\ref{supp:realbsr}, we include more details, as well as qualitative results, on the RealBSR-RAW dataset.

\section{Detailed Derivation of Robust Inverse ISP}
\label{supp:sec:invisp}
\subsection{ISP Definition}
Let $l\in[0,1]^3$ denote the linear RGB value of a pixel after black-level subtraction and normalization, written as 
\begin{equation}
l =
\begin{bmatrix}
l_R, l_G, l_B
\end{bmatrix}^T.
\end{equation}
The ISP function, $F_{ISP}$, maps $l$ to an sRGB output, $s\in[0,1]^3$, through the following four steps ~\cite{brooks2019unprocessing}.

\subsubsection{Step 1: White balancing.} Let 
\begin{equation}
W = \text{diag}(w_R, w_G, w_B)
\end{equation}
be the white-balancing gain matrix where $\text{diag}$ denotes the diagonal matrix. Define
\begin{equation}
u^{pre} = Wx, \quad u = \text{clip}(u^{pre}, 0, 1),
\end{equation}
where $\text{clip}$ represents truncating the values within $[0, 1]$.

\subsubsection{Step 2: Color correction.} Let $C\in \mathbb{R}^{3\times 3}$ denotes the color correction matrix (CCM). Then
\begin{equation}
v = Cu.
\end{equation}

\subsubsection{Step 3: Gamma compression.} Let $\alpha=1/\gamma$ with $\gamma=2.2$ in our implementation. We apply a lower clamp before gamma:
\begin{equation}
g = \phi(v)=\max(v,\varepsilon)^{\alpha},
\end{equation}
where $\epsilon=10^{-8}.$

\subsubsection{Step 4: Tone mapping.} We use the cubic tone mapping
\begin{equation}
s^{pre} = \psi(g)=3g^2-2g^3, \quad s = \text{clip}(s^{pre}, 0, 1),
\end{equation}
where $s$ is the sRGB output of the ISP.

\subsection{Jacobian Matrix}

With each component of the $F_{ISP}$, we can calculate the Jacobian matrix $J\in \mathbb{R}^{3 \times 3}$ of each pixel for the $\Delta L_{fo}$ and $\Delta L_{TSVD}$ in Eqn. 4 and 5 by the chain rule with six terms
\begin{equation}
J = F_{ISP}'(l) = \frac{\partial s}{\partial l} = \frac{\partial s}{\partial s^{pre}}\frac{\partial s^{pre}}{\partial g}\frac{\partial g}{\partial v}\frac{\partial v}{\partial u}\frac{\partial u}{\partial u^{pre}}\frac{\partial u^{pre}}{\partial l}.
\end{equation}

\subsubsection{1st term}
\begin{equation}
\frac{\partial s_c}{\partial s^{pre}_c} = m_c^s = 
\begin{cases}
1, &0 \leq s_c^{pre} \leq 1,\\
0, & otherwise,
\end{cases}
\quad c \in \{R, G, B\}.
\end{equation}
Thus, 
\begin{equation}
\frac{\partial s}{\partial s^{pre}} = D_s = \text{diag}(m^s_R,m^s_G,m^s_B).
\end{equation}

\subsubsection{2nd term}

\begin{equation}
\frac{\partial s^{pre}_c}{\partial g_c} = 6g_c(1 - g_c), \quad c \in \{R, G, B\}
\end{equation}
Hence,
\begin{equation}
\frac{\partial s^{pre}}{\partial g} = D_t = \text{diag}(6g_R(1 - g_R),6g_G(1 - g_G),6g_B(1 - g_B)).
\end{equation}

\subsubsection{3rd term}
\begin{equation}
\frac{\partial g_c}{\partial v_c} = m_c^\gamma\alpha\tilde{v}_c^{\alpha - 1}, \quad c \in \{R, G, B\},
\end{equation}
where 
\begin{equation}
\tilde{v}_c = \max(v_c, \epsilon), \quad
m_c^\gamma = 
\begin{cases}
1, & v_c \geq \epsilon,\\
0, & v_c < \epsilon,
\end{cases}
\end{equation}
Therefore,
\begin{equation}
\frac{\partial g}{\partial v} = D_\gamma =\text{diag}(m_R^\gamma\alpha\tilde{v}_R^{\alpha - 1},m_R^\gamma\alpha\tilde{v}_R^{\alpha - 1},m_R^\gamma\alpha\tilde{v}_R^{\alpha - 1}).
\end{equation}

\subsubsection{4th term}
\begin{equation}
\frac{\partial v}{\partial u} = C.
\end{equation}

\subsubsection{5th term}
\begin{equation}
\frac{\partial u_c}{\partial u^{pre}_c} = m_c^w = 
\begin{cases}
1, &0 \leq u_c^{pre} \leq 1,\\
0, & otherwise,
\end{cases}
\quad c \in \{R, G, B\}.
\end{equation}
Thus,
\begin{equation}
\frac{\partial u}{\partial u^{pre}} = D_w = \text{diag}(m_R^w, m_G^w, m_B^w).
\end{equation}

\subsubsection{6th term}
\begin{equation}
\frac{\partial u^{pre}}{\partial l} = W.
\end{equation}

\subsubsection{Final Jacobian Matrix}
Final Jacobian matrix is the combination of all six terms
\begin{equation}
J = D_sD_tD_\gamma C D_w W.
\end{equation}

\subsection{Mathematical Analyses}
In Eqn.~2 and 3 of the main paper, we approximate $\Delta L$ with first-order Taylor series expansion. The corresponding Taylor remainder term, representing the error introduced by this approximation, is defined as 
\begin{equation}
r(\Delta L) = F_{ISP}(L_b + \Delta L) - F_{ISP}(L_b) - F'_{ISP}(L_b)\Delta L,
\end{equation}
where $r(\Delta L)$ satisfies $||r(\Delta L)|| \approx 0$, ensuring its negligible impact on the approximation accuracy.

\subsubsection{Theorem.} Let $F_{ISP}: \mathbb{R}^3 \rightarrow \mathbb{R}^3$ be a per-pixel ISP function with a bounded Jacobian, $F'_{ISP}$, in a neighborhood of $L_b$. Then, if $F'_{ISP}$ is Lipschitz continuous with constant $U$, we have 
\begin{equation}
||r(\Delta L)|| \leq \frac{U}{2}||\Delta L||^2.
\end{equation}

\subsubsection{Proof.} By the fundamental theorem of calculus, we have 
\begin{equation}
F_{ISP}(L_b + \Delta L) - F_{ISP}(L_b) = \int_0^1 F'_{ISP}(L_b + t\Delta L)\Delta L dt.
\end{equation}
Thus,
\begin{equation}
\begin{aligned}
r(\Delta L) &= F_{ISP}(L_b + \Delta L) - F_{ISP}(L_b) - F'_{ISP}(L_b)\Delta L \\
&= \int_0^1 (F'_{ISP}(L_b + t\Delta L) - F'_{ISP}(L_b))\Delta L dt.
\end{aligned}
\end{equation}
Considering the components of ISP, where white balancing and color correction are linear operations, while gamma compression and tone mapping are bounded nonlinear transformations after clipping, we can assume that the Jacobian matrix $F'_{ISP}$ is Lipschitz continuous in a neighborhood of $L_b$, i.e., there exists $U > 0$ such that
\begin{equation}
|| F'_{ISP}(L_b + t\Delta L) - F'_{ISP}(L_b) || \leq U||t\Delta L||.
\end{equation}
Hence,
\begin{equation}
\begin{aligned}
||r(\Delta L)|| &= ||\int_0^1 (F'_{ISP}(L_b + t\Delta L) - F'_{ISP}(L_b))\Delta L dt ||\\
&\leq \int_0^1 ||F'_{ISP}(L_b + t\Delta L) - F'_{ISP}(L_b)||\:||\Delta L|| dt \\
&\leq \int_0^1 Ut||\Delta L||^2dt \\
&=\frac{U}{2}||\Delta L||^2. 
\end{aligned}
\end{equation}
\qed
In our experiments, $||\Delta L|| \ll 1$ with the 95th percentile measuring $1.61\times 10^{-2}$ on the SyntheticBurst dataset and $3.86\times 10^{-3}$ on real data. 
Therefore, we expect the residual error, bounded via $||r(\Delta L)|| \leq \frac{U}{2}||\Delta L||^2$, to be extremely small. 

\begin{algorithm}[!pbht]
	\DontPrintSemicolon
	\SetAlgoLined
	
	\KwInput{Base BISR linear restoration $L_b$,\\
	\,\,\,\,\,\,\,\,\,\,\,\,\,\,\,\,\,\,\,\,\,diffusion sRGB output $S_d$, \\
    \,\,\,\,\,\,\,\,\,\,\,\,\,\,\,\,\,\,\,\,\,ISP function $F_{ISP}$, \\
    \,\,\,\,\,\,\,\,\,\,\,\,\,\,\,\,\,\,\,\,\,regularization weight $\beta = 10^{-6}$,\\
    \,\,\,\,\,\,\,\,\,\,\,\,\,\,\,\,\,\,\,\,\,residual scale factor $\lambda_r = 1.0$,\\
    \,\,\,\,\,\,\,\,\,\,\,\,\,\,\,\,\,\,\,\,\,singular-value threshold $\epsilon = 1.0$.}
	\KwOutput{Inverted linear RGB $L_d$.}

    Compute BISR restoration rendering:
    \[
        S_b \leftarrow F_{ISP}(L_b).
    \]

    Compute diffusion residual:
    \[
        \Delta S \leftarrow S_d - S_b.
    \]

    Compute local ISP Jacobian at $L_b$:
    \[
    J \leftarrow F'_{ISP}(L_b).
    \]

    Compute singular values for $J$ of each pixel:
    \[
    \sigma_1, \sigma_2, \sigma_3 = \text{svdval}(J), \quad \text{where}\;\sigma_1 > \sigma_2 > \sigma_3
    \]

    Construct the reliability masks:
    \[
    m = \mathbf{1}[\sigma_3 < \epsilon].
    \]

    1st stage -- Compute the first-order residual update for well-conditioned pixels ($m = 1$):
    \[
    \Delta L_{fo} \approx (J^\intercal J + \beta I)^{-1}J^\intercal\Delta S.
    \]

    2nd stage -- For pixels marked as ill-conditioned ($m = 0$), compute truncated SVD:
    \[
    J \approx U_k \Sigma_k V_k^\top,\quad \text{where}\;\sigma_i > \epsilon,\;i\in\{1\dots k\}.
    \]
    
    2nd stage -- Compute TSVD residual updates:
    \[
    \Delta L_{TSVD}
    \approx
    V_k(\Sigma_k^\top \Sigma_k + \beta I)^{-1}
    \Sigma_k^\top U_k^\top \Delta S.
    \]

	Fuse the two residual updates anchored at $L_b$:
    \[
    \Delta L = m\Delta L_{fo} + (1 - m)\Delta L_{TSVD}, \quad L_d = L_b + \lambda_r\Delta L.
    \]

    Return $L_d$.
	\caption{Robust Inverse ISP}
	\label{alg:inv_isp}
\end{algorithm}

\subsection{Algorithm}
We demonstrate the complete algorithm of our robust inverse ISP in Alg.~\ref{alg:inv_isp}.

\section{Remark}
\label{supp:sec:remark}
\subsubsection{Why not perform end-to-end fine-tuning of the inverse ISP outputs in linear RGB space rather than fine-tuning the diffusion model in sRGB space?} We observe that back-propagating through the ISP pipeline introduces significant numerical instability, particularly when the initial input-output discrepancy is large. This instability is exacerbated by the behavior of linear solvers (e.g., SVD) in PyTorch during the optimization, where reliable gradients are critical.  Our empirical analysis indicates that end-to-end training fails to converge due to: (i) gradient explosion at luminance extremes (near-zero or saturated values), and (ii) non-differentiable clipping artifacts that disrupt the optimization manifold. Furthermore, prioritizing linear-space metrics through a complex ISP often leads to a degradation in perceptual sRGB quality, as the model overfits to sensor noise or quantization errors. This challenge is especially pronounced on BurstSR dataset, where supervision signals saturate in linear-space (PSNR $\sim$50 dB), providing negligible information for weight updates. In contrast, our approach ensures training stability and competitive fidelity without the overhead of differentiable camera pipelines~\cite{li2024dualdn,yu2021reconfigisp}, which we defer to future investigation.

\begin{table}[t]
	\caption{Quantitative evaluations on the SyntheticBurst dataset. \textbf{+Diffusion} represents the evaluation of diffusion outputs, and \textbf{+BurstGP(ours)} represents the evaluation of ISP re-rendered final outputs. Both are in sRGB space.}
	\footnotesize
	\centering
	\begin{adjustbox}{width=0.9\linewidth}
\begin{tabular}{lccccc}
\toprule
Method         & PSNR$~\uparrow$ & SSIM$~\uparrow$ & LPIPS$~\downarrow$ & TOPIQ$~\uparrow$ & MUSIQ$~\uparrow$\\ 
\midrule
\rowcolor{purple!10}BIPNet~\cite{dudhane2022burst} & 35.42 & \underline{0.9392} & 0.2270 & 0.8443 & 43.72  \\
\rowcolor{purple!10} \textbf{+Diffusion} & 33.43 & 0.9162 & 0.1705 & 0.8751 & 45.55 \\
\rowcolor{purple!10} \textbf{+BurstGP(ours)} & 33.43 & 0.9157 & 0.1712 & 0.8749 & 45.52 \\

\rowcolor{cyan!15}BurstM~\cite{kang2024burstm} & \underline{35.63} & 0.9383 & 0.2293 & 0.8437 & 44.66  \\
\rowcolor{cyan!15} \textbf{+Diffusion} & 34.05 & 0.9166 & 0.1615 & 0.8835  & 45.62 \\
\rowcolor{cyan!15} \textbf{+BurstGP(ours)} & 34.02 & 0.9165 & 0.1618 & 0.8836  & 45.57  \\

\rowcolor{blue!15}BSRT-S~\cite{luo2022bsrt} & 35.51 & 0.9391 & 0.2275 & 0.8473 & 44.48  \\
\rowcolor{blue!15} \textbf{+Diffusion} & 33.99 & 0.9160 & 0.1629 & \underline{0.8853} & 45.75  \\
\rowcolor{blue!15} \textbf{+BurstGP(ours)} & 33.95 & 0.9158 & 0.1634 & \underline{0.8853} & 45.70  \\

\rowcolor{violet!15}BSRT-L~\cite{luo2022bsrt} & \textbf{36.55} & \textbf{0.9462} & 0.2114 & 0.8656 & 45.69  \\
\rowcolor{violet!15} \textbf{+Diffusion} & 34.30 & 0.9196 & \textbf{0.1542} & \textbf{0.8913} & \textbf{46.75} \\
\rowcolor{violet!15} \textbf{+BurstGP(ours)} & 34.29 & 0.9194 & \underline{0.1543} & \textbf{0.8913} & \underline{46.72} \\

\bottomrule
\end{tabular}
	\end{adjustbox}
	\label{tab:syn_diff_quan}
\end{table}
\begin{table}[t]
	\caption{
        Quantitative evaluations on the BurstSR (real) dataset.
        Following prior work~\cite{luo2022bsrt}, all full-reference metrics are computed post-alignment, except ST-LPIPS. \textbf{+Diffusion} represents the evaluation of diffusion outputs, and \textbf{+BurstGP(ours)} represents the evaluation of ISP re-rendered final outputs. Both are in sRGB space.
    }
	\footnotesize
	\centering
	\begin{adjustbox}{width=\linewidth}
\begin{tabular}{lcccccccc}
\toprule
Method         & PSNR$~\uparrow$ & SSIM$~\uparrow$ & LPIPS$~\downarrow$ & ST-LPIPS$~\downarrow$ & TOPIQ$~\uparrow$ & MUSIQ$~\uparrow$\\ 
\midrule
\rowcolor{purple!10}BIPNet~\cite{dudhane2022burst} & 29.91 & 0.8924 & 0.3337 & 0.2289 & 0.6380 & 38.42 \\
\rowcolor{purple!10}\textbf{+Diffusion} & 30.61 & 0.9010 & 0.3060 & 0.2095 & 0.6981 & 43.65  \\
\rowcolor{purple!10}\textbf{+BurstGP(ours)} & 30.61 & 0.9010 & 0.3059 & 0.2101 & 0.6980 & 43.65  \\
\rowcolor{cyan!15}BurstM~\cite{kang2024burstm} & 30.31 & 0.8989 & 0.3231 & 0.2076 & 0.6725 & 43.57 \\
\rowcolor{cyan!15}\textbf{+Diffusion} & 30.84 & 0.9052 & 0.3000 & 0.1954 & 0.7103 & 46.49  \\
\rowcolor{cyan!15}\textbf{+BurstGP(ours)} & 30.88 & 0.9052 & 0.3000 & 0.1962 & 0.7105 & 46.48  \\
\rowcolor{blue!15}BSRT-S~\cite{luo2022bsrt} & 30.53 & 0.9034 & 0.3190 & 0.2043 & 0.6873 & 43.16 \\
\rowcolor{blue!15}\textbf{+Diffusion} & \underline{30.98}  & \underline{0.9078} & 0.2961 & 0.1897 & 0.7241 & 47.26 \\
\rowcolor{blue!15}\textbf{+BurstGP(ours)} & \underline{30.98}  & \textbf{0.9080} & 0.2976 & 0.1903 & 0.7238 & 47.28 \\
\rowcolor{violet!15}BSRT-L~\cite{luo2022bsrt} & 30.59 & 0.9050 & 0.3111 & 0.1981 & 0.6976 & 44.44 \\
\rowcolor{violet!15}\textbf{+Diffusion} & \textbf{30.99} & \textbf{0.9080} & \textbf{0.2944} & \textbf{0.1873} & \textbf{0.7269} & \underline{47.86} \\
\rowcolor{violet!15}\textbf{+BurstGP(ours)} & \textbf{30.99} & \textbf{0.9080} & \underline{0.2960} & \underline{0.1881} & \underline{0.7267} & \textbf{47.89} \\
\bottomrule
\end{tabular}
	\end{adjustbox}
	\label{tab:bsr_diff_quan}
\end{table}

\begin{figure}[tb]
  \centering
  \begin{minipage}{\linewidth}
    \includegraphics[width=\linewidth]{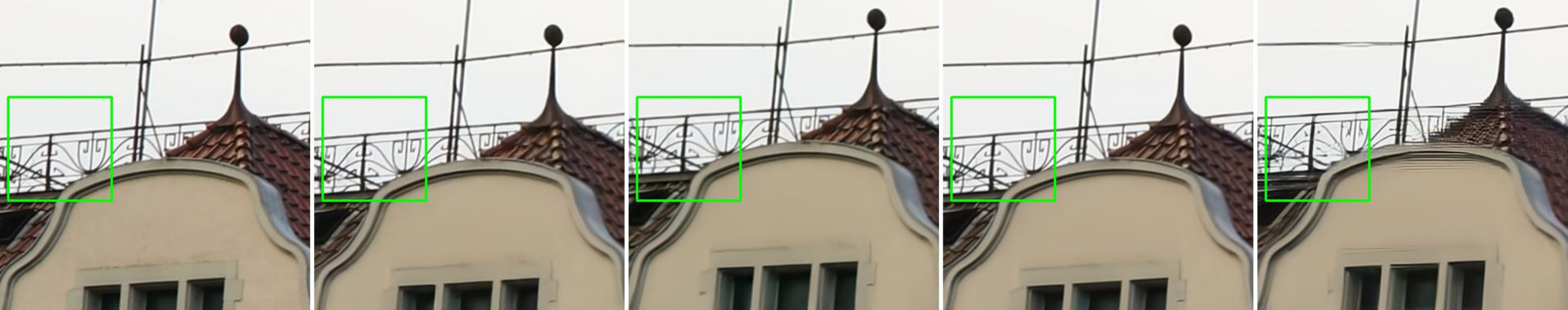}
    \vfill \vspace*{0.01cm}
  \end{minipage}
  \vfill
  \begin{minipage}{\linewidth}
    \includegraphics[width=\linewidth]{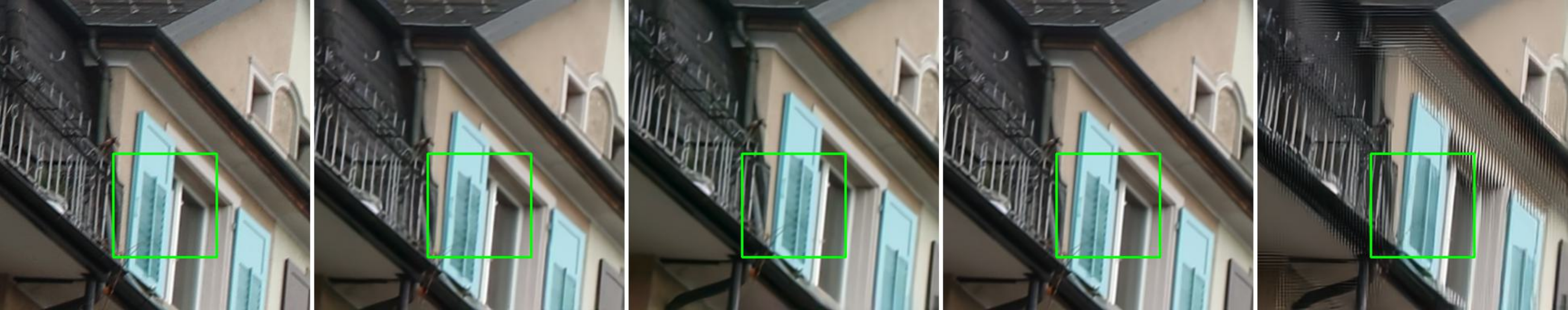}
    \vfill \vspace*{0.01cm}
  \end{minipage}
  \vfill
  \begin{minipage}{\linewidth}
    \includegraphics[width=\linewidth]{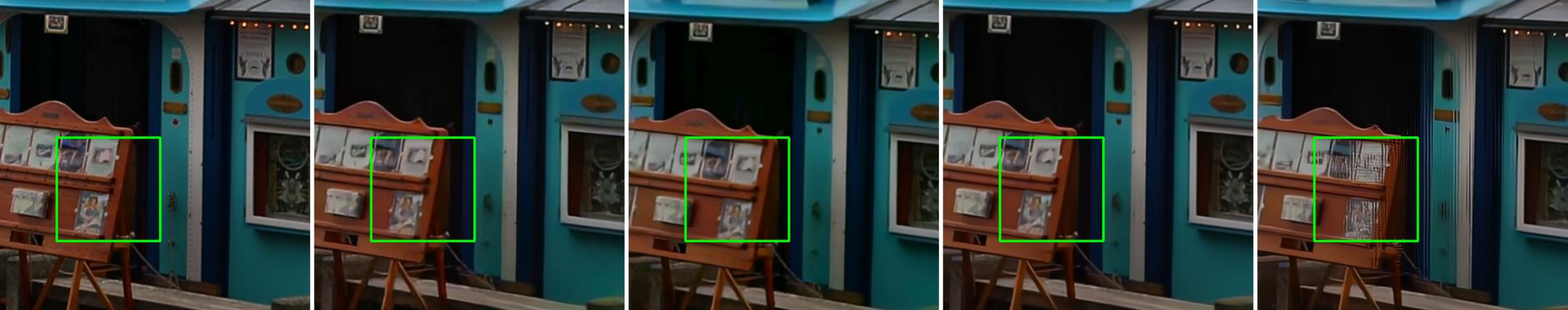}
    \vfill \vspace*{-0.12cm}
    \text{\scriptsize ~~~~~~~~~~GT~~~~~~~~~~~~w/ permutation training (ref, non-ref)~w/o permutation training (ref, non-ref)}
    \vfill \vspace*{0.12cm}
  \end{minipage}
  \caption{Qualitative results on the SyntheticBurst dataset with/without permutation training. Without permutation training, the BISR model is unable to recover details in non-reference frames, even those it is able to recover for the reference frame (first two examples). The model also produces artifacts in non-reference frames without permutation training (last example). Permutation training solves these issues.
  }
  \label{fig:syn_permutation_supp}
\end{figure}

\begin{figure}[tb]
  \centering
  \begin{minipage}{\linewidth}
    \includegraphics[width=\linewidth]{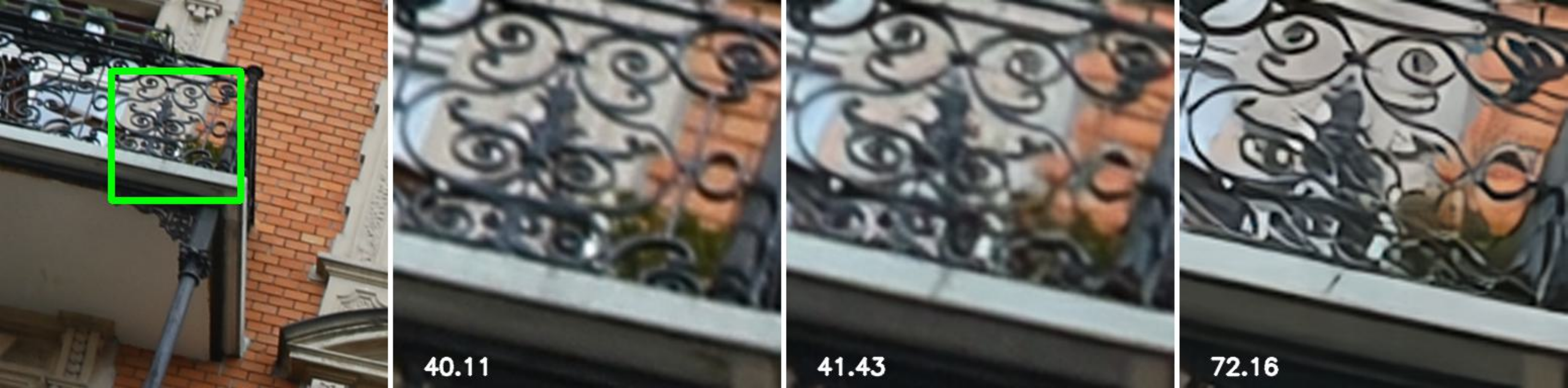}
    \vfill \vspace*{0.01cm}
  \end{minipage}
  \vfill
  \begin{minipage}{\linewidth}
    \includegraphics[width=\linewidth]{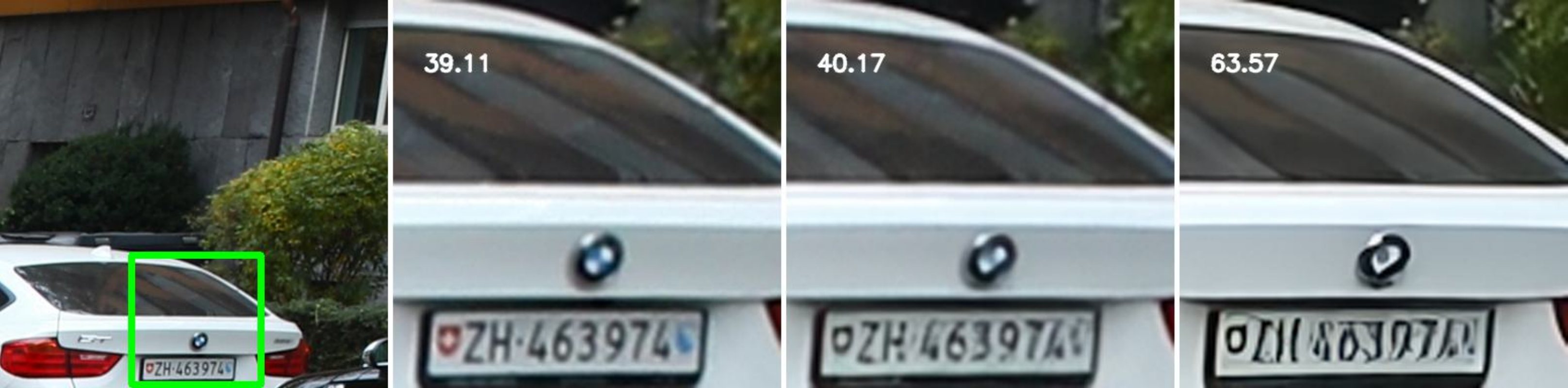}
    \vfill \vspace*{0.01cm}
  \end{minipage}
  \vfill
  \begin{minipage}{\linewidth}
    \includegraphics[width=\linewidth]{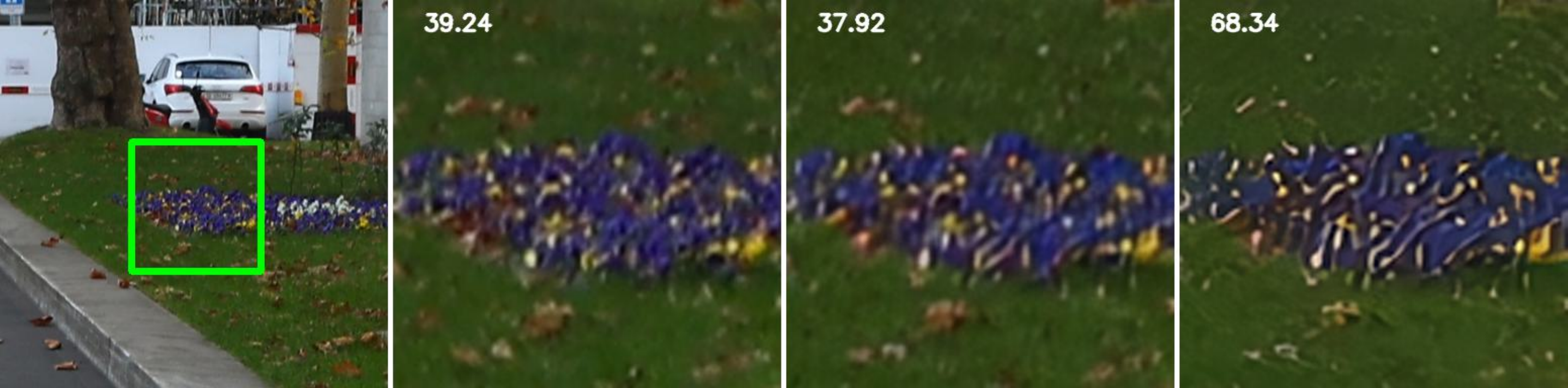}
    \vfill \vspace*{-0.12cm}
    \text{\scriptsize ~~~~~~~~~~~~~~GT~~~~~~~~~~~~~~~~~~~~~~~~~~~~GT~~~~~~~~~~~~~~~~~~~~with finetuning~~~~~~~~~~without finetuning}
    \vfill \vspace*{0.12cm}
  \end{minipage}
  \caption{Qualitative results on the SyntheticBurst dataset with/without finetuning diffusion model. We use BSRT-L as the BISR model. We report MUSIQ for the entire image. We observe severe artifacts when diffusion model is not finetuned, which we correct in BurstGP by finetuning the diffusion model on burst dataset.
  }
  \label{fig:syn_finetuning_supp}
\end{figure}
\begin{figure}[tb]
  \centering
  \begin{minipage}{\linewidth}
    \includegraphics[width=\linewidth]{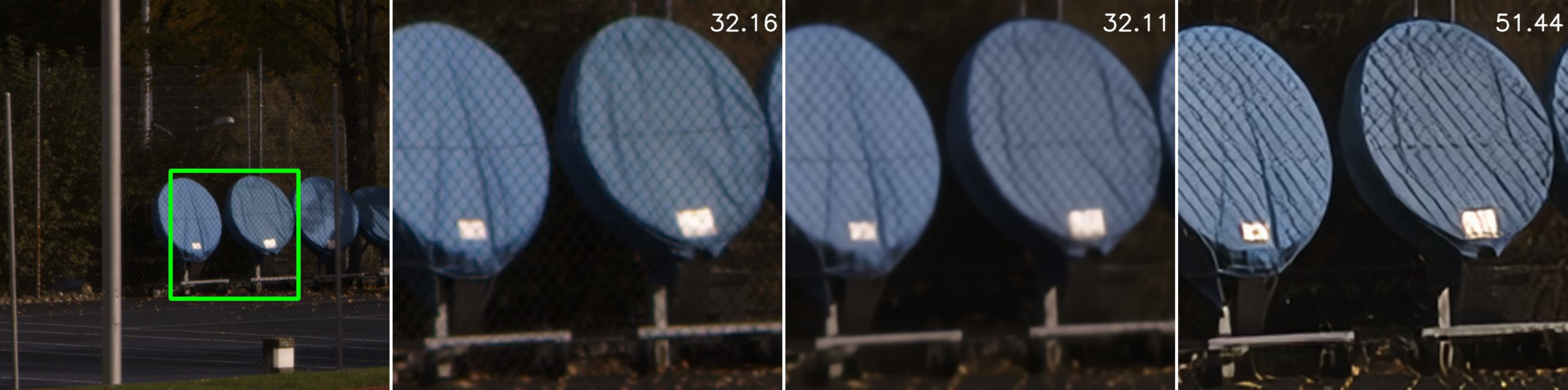}
    \vfill \vspace*{0.01cm}
  \end{minipage}
  \vfill
  \begin{minipage}{\linewidth}
    \includegraphics[width=\linewidth]{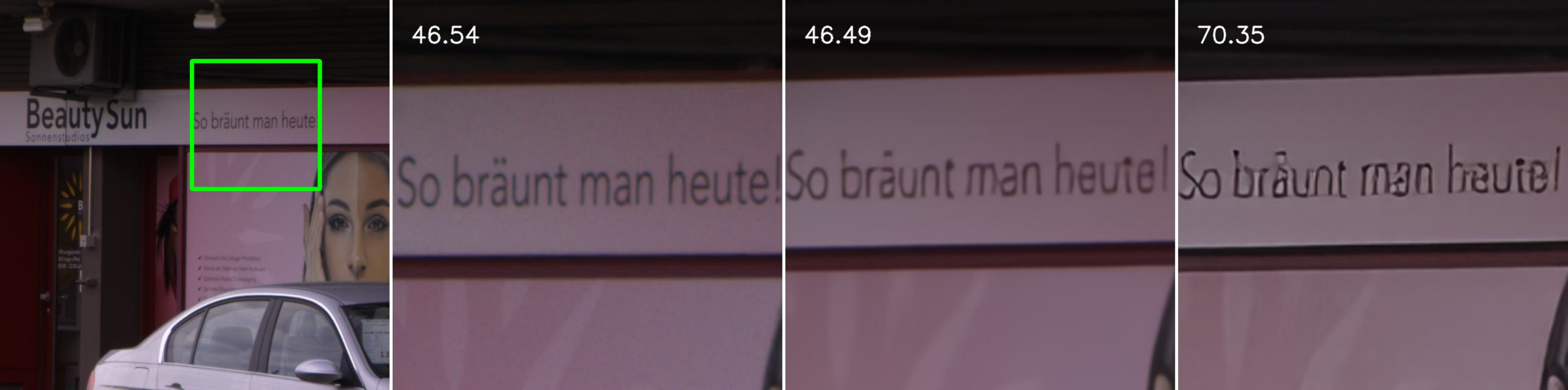}
    \vfill \vspace*{0.01cm}
  \end{minipage}
  \vfill
  \begin{minipage}{\linewidth}
    \includegraphics[width=\linewidth]{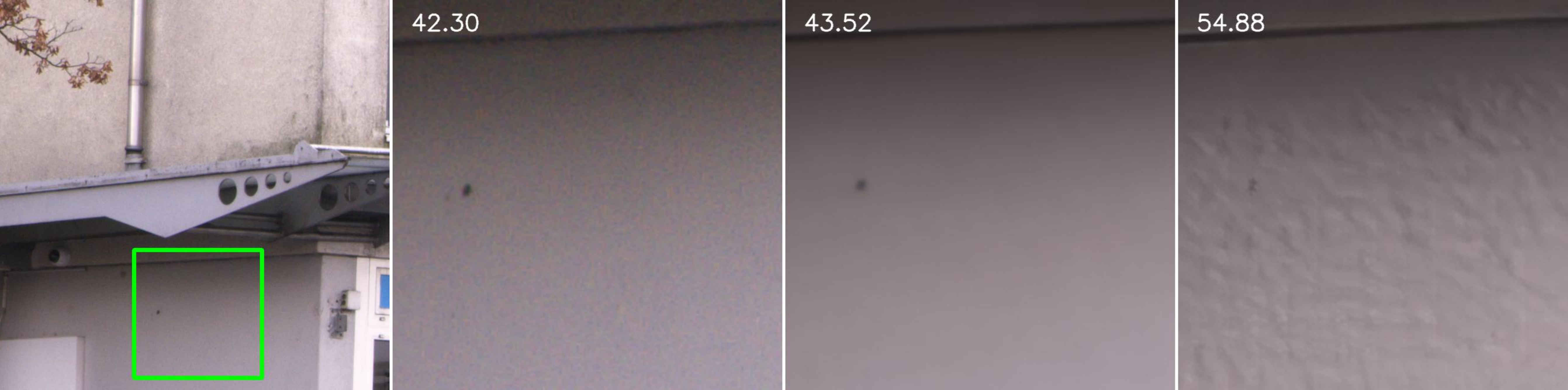}
    \vfill \vspace*{-0.12cm}
    \text{\scriptsize ~~~~~~~~~~~~~~GT~~~~~~~~~~~~~~~~~~~~~~~~~~~~GT~~~~~~~~~~~~~~~~~~~~with finetuning~~~~~~~~~~without finetuning}
    \vfill \vspace*{0.12cm}
  \end{minipage}
  \caption{Qualitative results on the BurstSR dataset with/without finetuning the diffusion model. We use BSRT-L as the BISR model. We report MUSIQ for the entire image. We observe severe artifacts when the diffusion model is not finetuned, which we correct in BurstGP by finetuning the diffusion model on a burst dataset.
  }
  \label{fig:burstsr_finetuning_supp}
\end{figure}

\section{Additional Results}
\label{supp:sec:newresults}

\subsection{Evaluations of Diffusion Outputs}
We assess the performance of our fine-tuned diffusion model before the robust inverse ISP in Tab.~\ref{tab:syn_diff_quan} and Tab.~\ref{tab:bsr_diff_quan}. Specifically, \textbf{BISR + Diffusion} evaluates the direct diffusion outputs in sRGB space, while \textbf{BISR + BurstGP(ours)} evaluates the final outputs after ISP re-rendering, where linear-space results are converted back to sRGB. Note that both scenarios utilize our fine-tuned model, with the only difference being the processing of the output. As expected, the diffusion outputs exhibit marginally superior quality compared to the re-rendered results, reflecting the inherent trade-off between perceptual enhancement and physical consistency introduced by the ISP inversion process.

\subsection{BISR outputs without permutation training}
In Fig. \ref{fig:syn_permutation_supp}, we show qualitative results from a BISR when trained with and without permutation training on the SyntheticBurst dataset. As expected, BISR model produces faithful reconstruction of the reference frame with or without permutation training. However, note that without permutation training, BISR model fails to reconstruct the same details on non-reference frame that were already well-constructed in the reference frame. Without permutation training, BISR model also produces artifacts on non-reference frame. We solve both challenges with permutation training, which results in temporally-consistent and artifact-free reconstructions for all frames. See also Tab.~6 of the main paper for the quantitative impact of ablating permutation training, which includes a severe loss in fidelity. In particular, enabling permutation training in the base BISR model improves the performance of BurstGP by 3.26 dB.

\subsection{Results without finetuning}
In Fig. \ref{fig:syn_finetuning_supp} and \ref{fig:burstsr_finetuning_supp}, we present visual comparison of the final SR outputs when the diffusion model was or was not finetuned on the burst dataset (i.e., our model vs.\ the initial pretrained DOVE). While we observe higher MUSIQ without any additional finetuning on the burst dataset, the synthetic details are often hallucinatory (e.g., distortions on regions with plants). We also observe jarring over-enhancement on certain regions, such as text (e.g., vehicle number plates), uniform areas (e.g., walls), and regular patterns (e.g., nets). Finetuning the diffusion model on burst datasets overcomes the above challenges and produces results that are more faithful to the GT. Tab. 6 of the main paper shows that BurstGP without finetuning on burst datasets exhibits significantly reduced fidelity, with a performance degradation of 5.40 dB compared to the fine-tuned version.

\begin{figure}[t]
    \centering
    \includegraphics[width=0.99\textwidth]{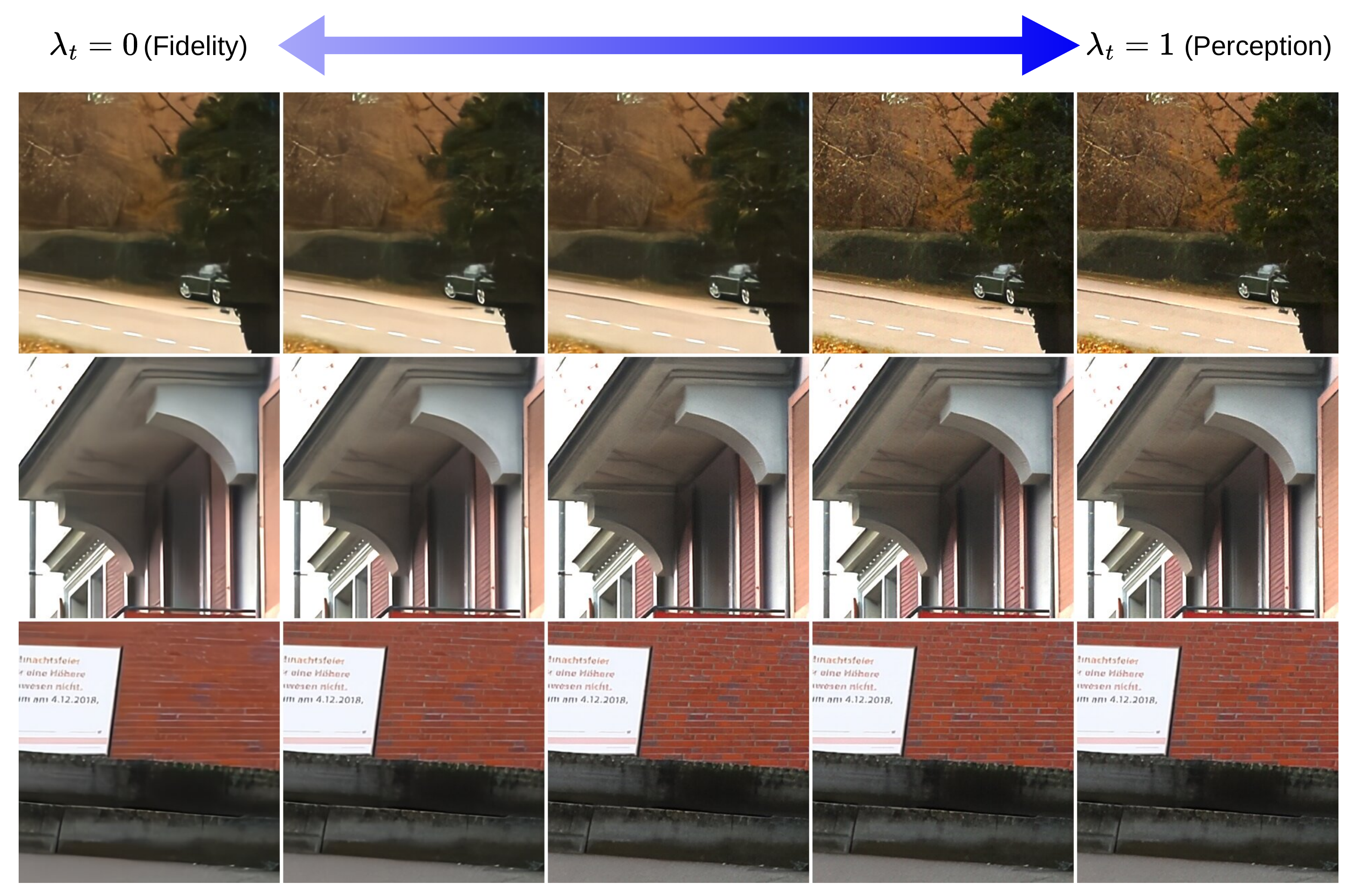}
    \caption{Qualitative results from BurstGP with BurstM, from high fidelity (low $\lambda_t$) \textit{(left)} to high perceptual quality (high $\lambda_t$) \textit{(right)}. Similar to $\lambda_r$, we observe that low $\lambda_t$ stays closer to the base BISR model output, which has less detail and is slightly oversmoothed, while high $\lambda_t$ introduces additional sharp texture to the image, as the network assumes additional image content needs to be generated. (zoom-in for better view)}
    \label{fig:lambda_t}
\end{figure}

\subsection{Analyses of degradation conditioning and $\lambda_t$}
Our diffusion backbone employs a distilled one-step architecture that operates at a fixed timestep ($t = 399$) during both training and inference. This framework treats the encoded ISP-rendered BISR outputs as initial noisy latents, which are then processed through a single denoising step to generate the restored output.

Ideally, the diffusion model should behave conservatively when the BISR reconstruction is already perceptually plausible (i.e., when the degradation is small), while applying stronger generative refinement when the BISR result is suboptimal (i.e., when the degradation is large). A straightforward approach would be to manually adjust the timestep $t$: decreasing $t$ for easier cases to reduce denoising and increasing $t$ for harder cases to encourage stronger generative refinement. However, as shown in Tab.~\ref{tab:syn_timestep}, performance degrades significantly when using different timesteps. This occurs because the generative prior is distilled and optimized for a fixed timestep, making the model sensitive to deviations from this operating point.

To overcome this limitation, we inject a degradation-dependent embedding into the timestep pathway, enabling adaptive restoration strength while preserving the pretrained diffusion behavior. Experimental results in Tab.\ 6 of the main paper show that the degradation conditioning improves both fidelity and perceptual quality. To further analyze its effect, we perform a brute-force search over timestep embeddings ($t \in \{1, \dots, 1000\}$) and identify the closest timestep to the learned degradation-conditioned embedding. The results indicate that the learned embedding corresponds approximately to a timestep of $\mathbf{t \approx 390}$ on SyntheticBurst and $\mathbf{t \approx 371}$ on real data. In contrast, even a small deviation from the fixed timestep without degradation-aware conditioning leads to noticeable performance drops as shown in Tab.~\ref{tab:syn_timestep}. 

Although the primary goal of this mechanism is to modulate the effective timestep according to the degradation level, we observe that it also shifts the perception–distortion operating point as a secondary effect. In particular, adjusting $\lambda_t$ allows explicit control over the fidelity–perception trade-off, as illustrated in Fig.~\ref{fig:lambda_t} and Tab. 4(a) of the main paper, where decreasing $\lambda_t$ results in more conservative reconstructions.

\begin{figure}[tb]
  \centering
  \begin{minipage}{\linewidth}
    \includegraphics[width=\linewidth]{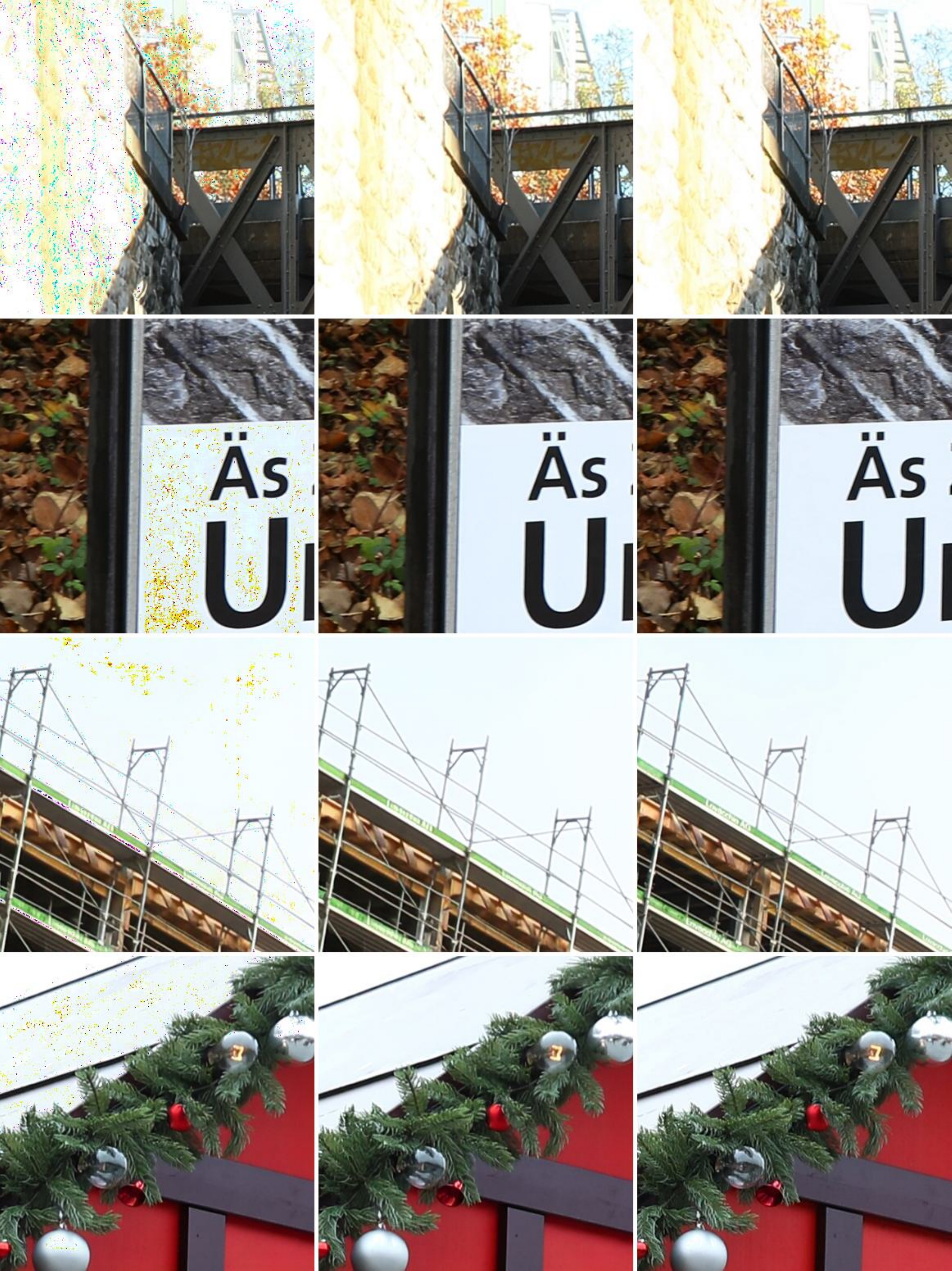}
    \vfill \vspace*{-0.12cm}
    \text{\scriptsize ~~~~~~~~~~~1st stage outputs~~~~~~~~~~~~~~~~~~~~2nd stage outputs~~~~~~~~~~~~~~~~~~~~~~~~~~~~GT~~~~~~~~~~~}
    \vfill \vspace*{0.12cm}
  \end{minipage}

  \caption{Qualitative results of two-stage correction scheme in the robust inverse ISP. We re-render the linear outputs back to sRGB space with the ISP operator for visualization. The initial first-order updates in the 1st stage exhibit instability for ill-conditioned pixels (particularly in saturated regions), leading to overflow-inducing residual estimates. The TSVD refinement in the 2nd stage effectively suppresses these overflow artifacts, ensuring physically plausible outputs. 
  }
  \label{fig:first_order}
\end{figure}

\subsection{Impact of Two-stage Inverse ISP}
Our robust inverse ISP method employs a two-stage correction scheme: (i) first-order updates for well-conditioned pixels, followed by (ii) low-rank truncated-SVD (TSVD) refinement to handle ill-conditioned (rank-deficient) pixels where the ISP Jacobian matrix becomes unstable. As shown in Fig.~\ref{fig:first_order}, the first-order stage alone produces overflow artifacts in ill-conditioned regions (manifesting as random noise), while the TSVD refinement effectively suppresses these artifacts through truncation of near-zero singular values in the Jacobian matrix. The presence of these overflow artifacts results in a significant PSNR degradation of 2.12 dB.

\subsection{More qualitative visualizations} 
\begin{figure}[tb]
  \centering
  \begin{minipage}{\linewidth}
    \includegraphics[width=\linewidth]{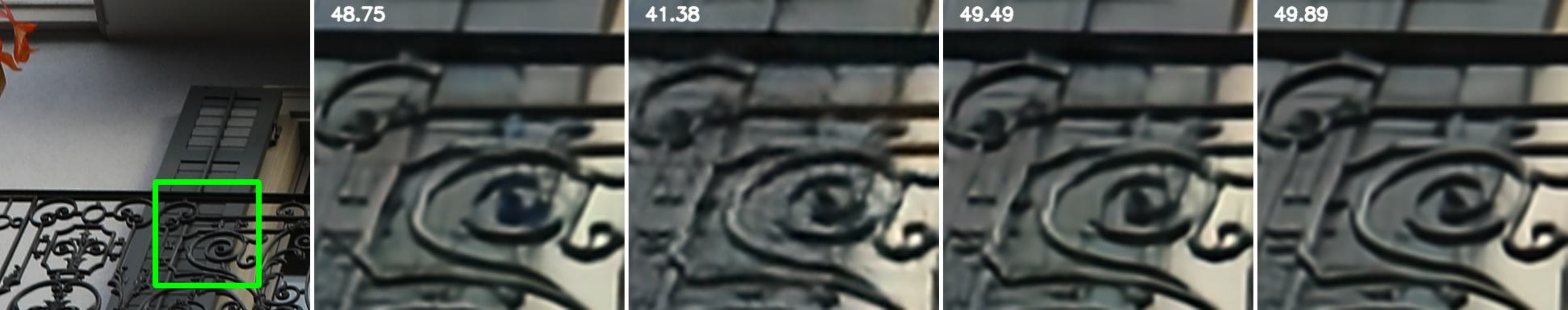}
    \vfill \vspace*{-0.12cm}
    \text{\scriptsize ~~~~~~~~~~GT~~~~~~~~~~~~~~~~~~~~BIPNet~~~~~~~~~~~~~~~~BurstM~~~~~~~~~~~~~~~BSRT-S~~~~~~~~~~~~~~~~BSRT-L}
    \vfill \vspace*{0.12cm}
  \end{minipage}
  \vfill
  \begin{minipage}{\linewidth}
    \includegraphics[width=\linewidth]{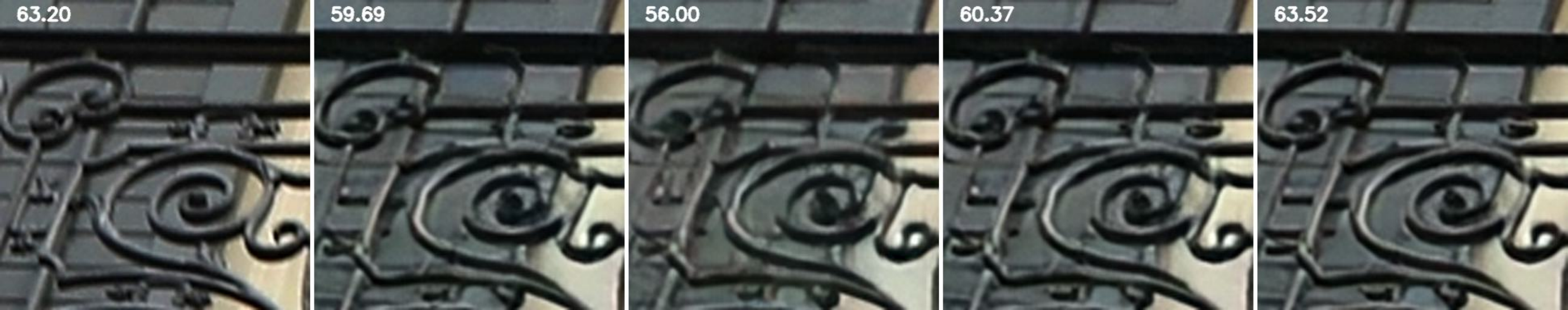}
    \vfill \vspace*{-0.12cm}
    \text{\scriptsize \hspace{2.45cm} BIPNet+BurstGP~~BurstM+BurstGP~~BSRT-S+BurstGP~~BSRT-L+BurstGP}
    \vfill \vspace*{0.12cm}
  \end{minipage}
  \vfill
  \begin{minipage}{\linewidth}
    \includegraphics[width=\linewidth]{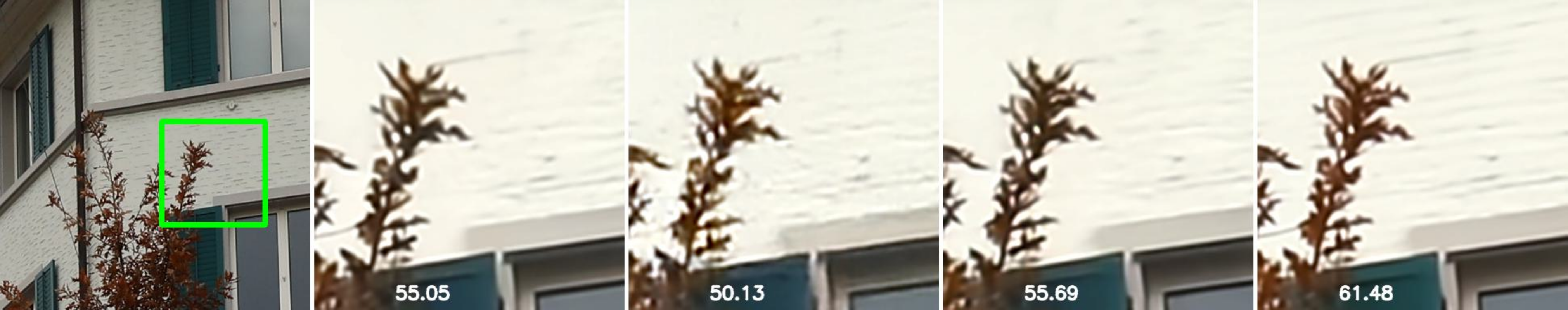}
    \vfill \vspace*{-0.12cm}
    \text{\scriptsize ~~~~~~~~~~GT~~~~~~~~~~~~~~~~~~~~BIPNet~~~~~~~~~~~~~~~~BurstM~~~~~~~~~~~~~~~BSRT-S~~~~~~~~~~~~~~~~BSRT-L}
    \vfill \vspace*{0.12cm}
  \end{minipage}
  \vfill
  \begin{minipage}{\linewidth}
    \includegraphics[width=\linewidth]{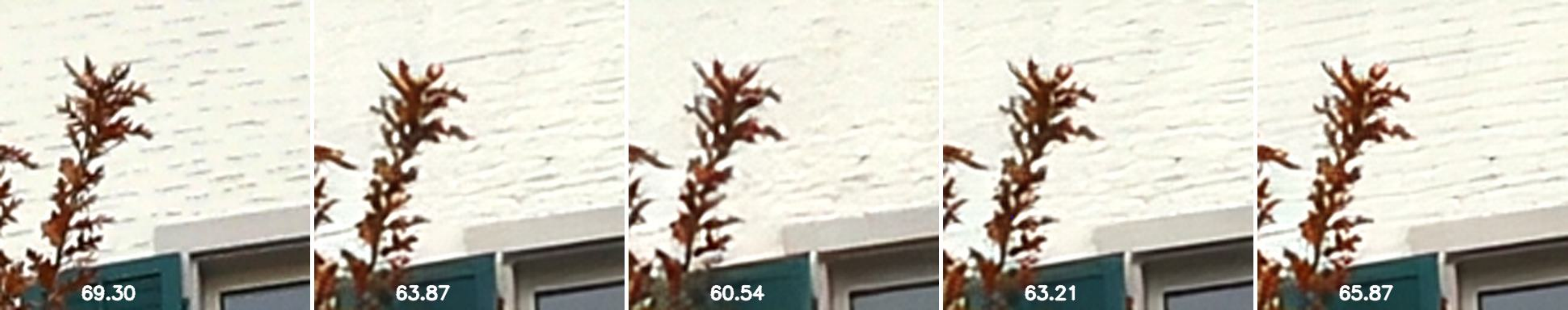}
    \vfill \vspace*{-0.12cm}
    \text{\scriptsize \hspace{2.45cm} BIPNet+BurstGP~~BurstM+BurstGP~~BSRT-S+BurstGP~~BSRT-L+BurstGP}
    \vfill \vspace*{0.12cm}
  \end{minipage}
  \vfill
  \begin{minipage}{\linewidth}
    \includegraphics[width=\linewidth]{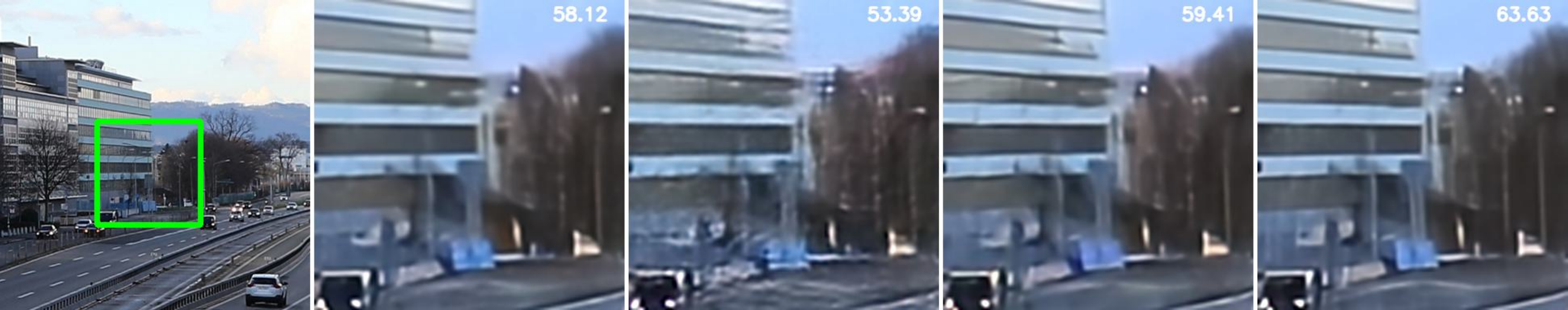}
    \vfill \vspace*{-0.12cm}
    \text{\scriptsize ~~~~~~~~~~GT~~~~~~~~~~~~~~~~~~~~BIPNet~~~~~~~~~~~~~~~~BurstM~~~~~~~~~~~~~~~BSRT-S~~~~~~~~~~~~~~~~BSRT-L}
    \vfill \vspace*{0.12cm}
  \end{minipage}
  \vfill
  \begin{minipage}{\linewidth}
    \includegraphics[width=\linewidth]{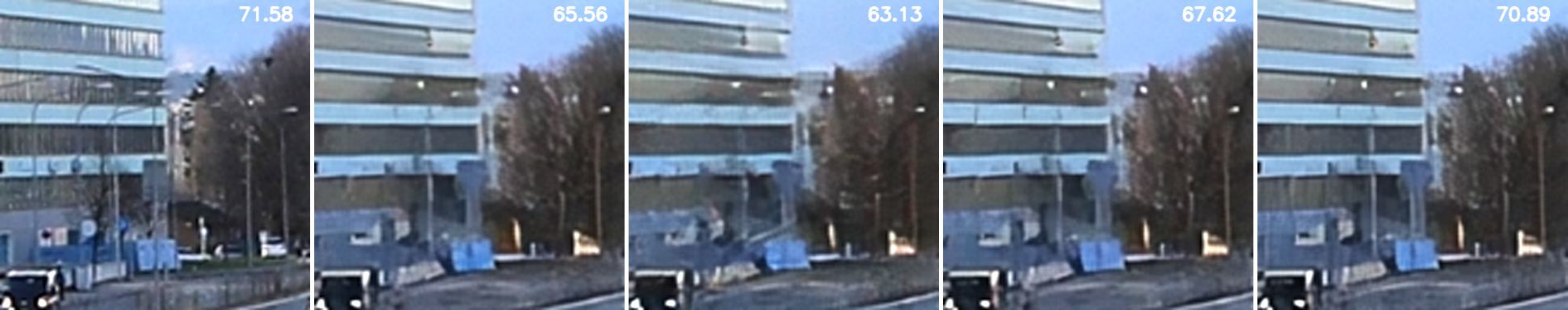}
    \vfill \vspace*{-0.12cm}
    \text{\scriptsize \hspace{2.45cm} BIPNet+BurstGP~~BurstM+BurstGP~~BSRT-S+BurstGP~~BSRT-L+BurstGP}
    \vfill \vspace*{0.12cm}
  \end{minipage}
  \caption{Qualitative results on the SyntheticBurst dataset. We report MUSIQ for the entire image. We brighten insets for all examples for better visibility.
  Across models, the addition of our BurstGP method is able to improve denoising and correct distortions (e.g., the first image set), generate plausible textures (e.g., the wall of the second set), and augment blurred content with new details (e.g., the tree in the third set).
  }
  \label{fig:syn_qual_supp}
\end{figure}

\begin{figure}[tb]
  \centering
  \begin{minipage}{\linewidth}
    \includegraphics[width=\linewidth]{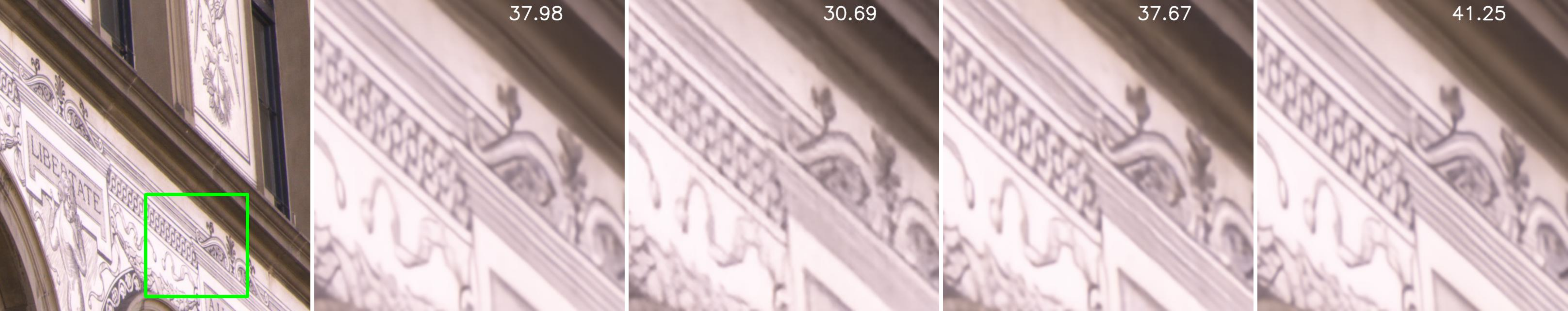}
    \vfill \vspace*{-0.12cm}
    \text{\scriptsize ~~~~~~~~~~GT~~~~~~~~~~~~~~~~~~~~BIPNet~~~~~~~~~~~~~~~~BurstM~~~~~~~~~~~~~~~BSRT-S~~~~~~~~~~~~~~~~BSRT-L}
    \vfill \vspace*{0.12cm}
  \end{minipage}
  \vfill
  \begin{minipage}{\linewidth}
    \includegraphics[width=\linewidth]{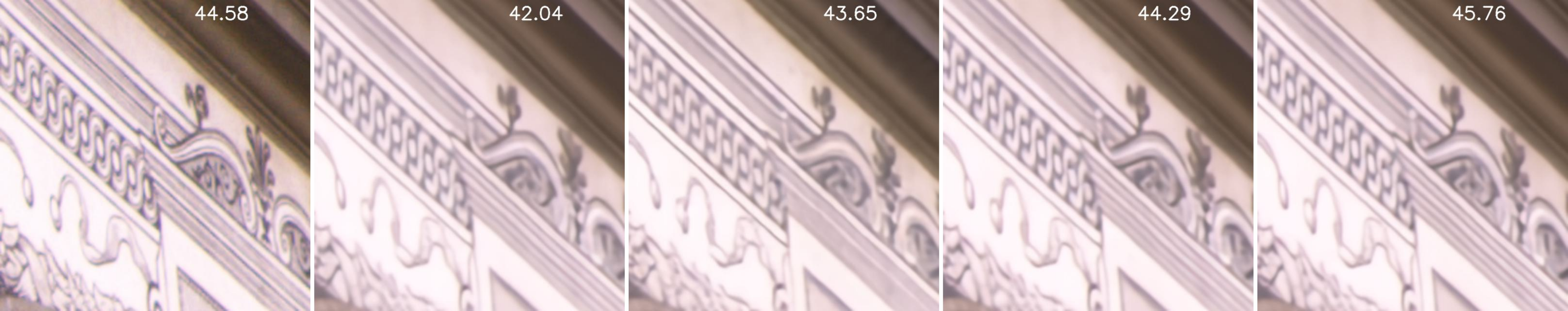}
    \vfill \vspace*{-0.12cm}
    \text{\scriptsize \hspace{2.45cm} BIPNet+BurstGP~~BurstM+BurstGP~~BSRT-S+BurstGP~~BSRT-L+BurstGP}
    \vfill \vspace*{0.12cm}
  \end{minipage}
  \vfill
  \begin{minipage}{\linewidth}
    \includegraphics[width=\linewidth]{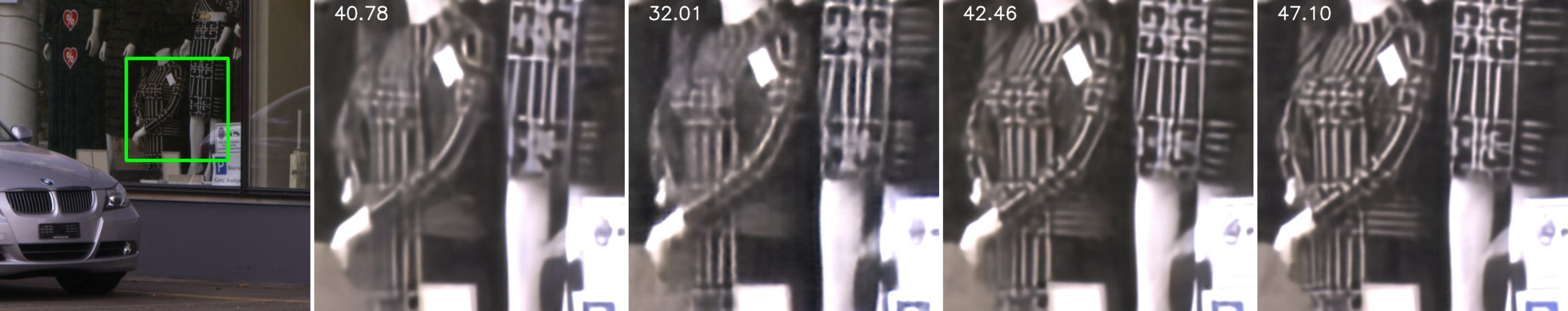}
    \vfill \vspace*{-0.12cm}
    \text{\scriptsize ~~~~~~~~~~GT~~~~~~~~~~~~~~~~~~~~BIPNet~~~~~~~~~~~~~~~~BurstM~~~~~~~~~~~~~~~BSRT-S~~~~~~~~~~~~~~~~BSRT-L}
    \vfill \vspace*{0.12cm}
  \end{minipage}
  \vfill
  \begin{minipage}{\linewidth}
    \includegraphics[width=\linewidth]{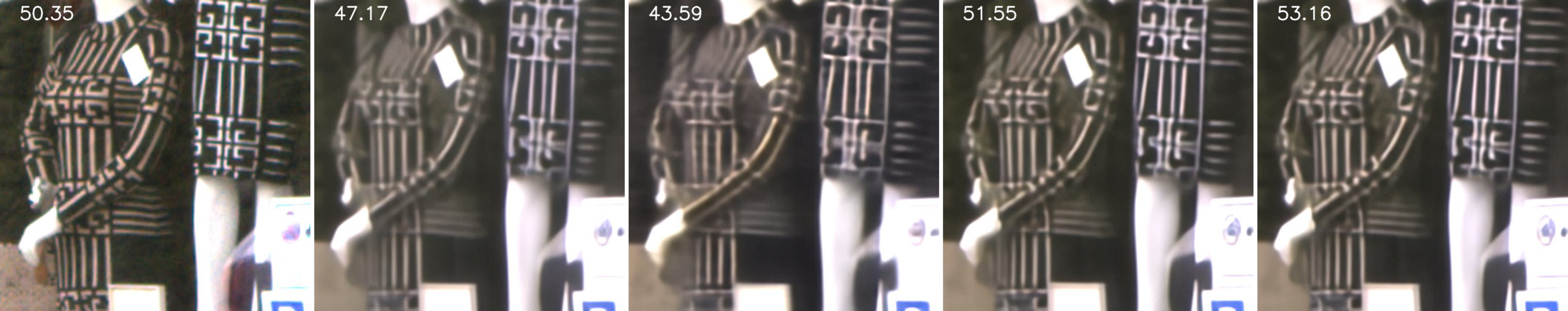}
    \vfill \vspace*{-0.12cm}
    \text{\scriptsize \hspace{2.45cm} BIPNet+BurstGP~~BurstM+BurstGP~~BSRT-S+BurstGP~~BSRT-L+BurstGP}
    \vfill \vspace*{0.12cm}
  \end{minipage}
  \vfill
  \begin{minipage}{\linewidth}
    \includegraphics[width=\linewidth]{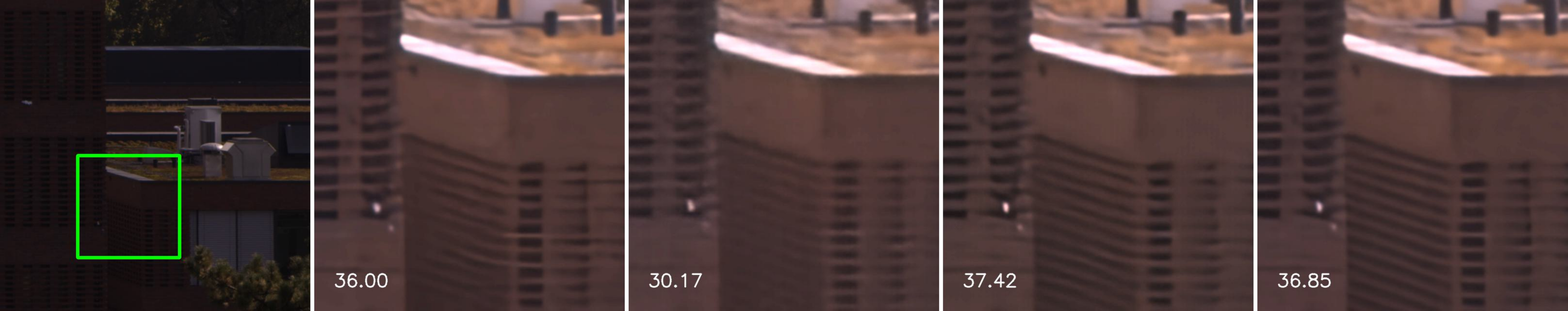}
    \vfill \vspace*{-0.12cm}
    \text{\scriptsize ~~~~~~~~~~GT~~~~~~~~~~~~~~~~~~~~BIPNet~~~~~~~~~~~~~~~~BurstM~~~~~~~~~~~~~~~BSRT-S~~~~~~~~~~~~~~~~BSRT-L}
    \vfill \vspace*{0.12cm}
  \end{minipage}
  \vfill
  \begin{minipage}{\linewidth}
    \includegraphics[width=\linewidth]{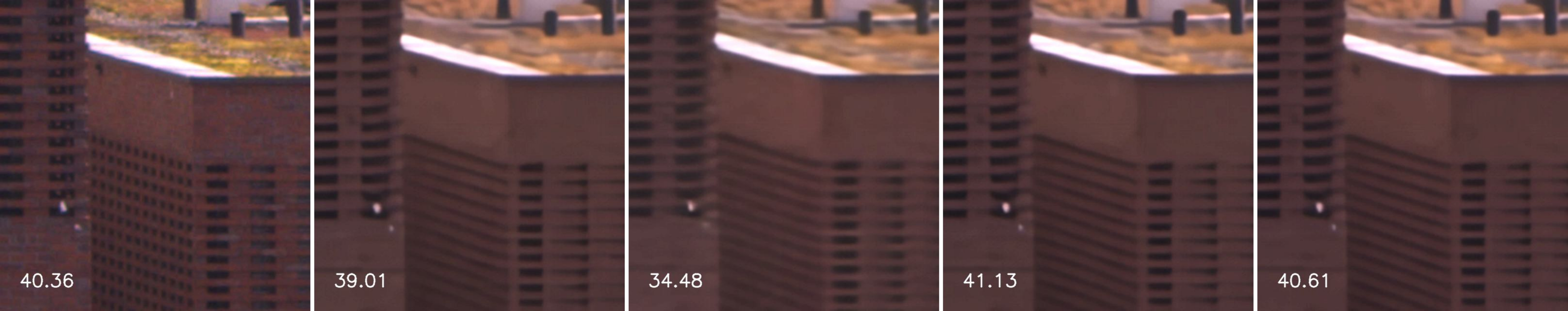}
    \vfill \vspace*{-0.12cm}
    \text{\scriptsize \hspace{2.45cm} BIPNet+BurstGP~~BurstM+BurstGP~~BSRT-S+BurstGP~~BSRT-L+BurstGP}
    \vfill \vspace*{0.12cm}
  \end{minipage}
  \caption{Qualitative results on BurstSR. We report MUSIQ for the entire image. We brighten the bottom two examples for better visibility.
  In all cases, BurstGP reduces blur and oversmoothing (e.g., the lines of the first image set).
  Further, it can mitigate artifacts from the base model, resolving real details closer to the GT (e.g., the lines on the clothes in the second set, as well as the blurred horizontal lines in the third). 
  }
  \label{fig:burstsr_qual_supp}
\end{figure}

We show additional qualitative results on the SyntheticBurst and BurstSR datasets in Fig.~\ref{fig:syn_qual_supp} and Fig.~\ref{fig:burstsr_qual_supp}, respectively. Our BurstGP consistently reconstructs results with higher quality than BISR counterparts. 


\begin{figure}
    \centering
    \includegraphics[width=0.995\linewidth]{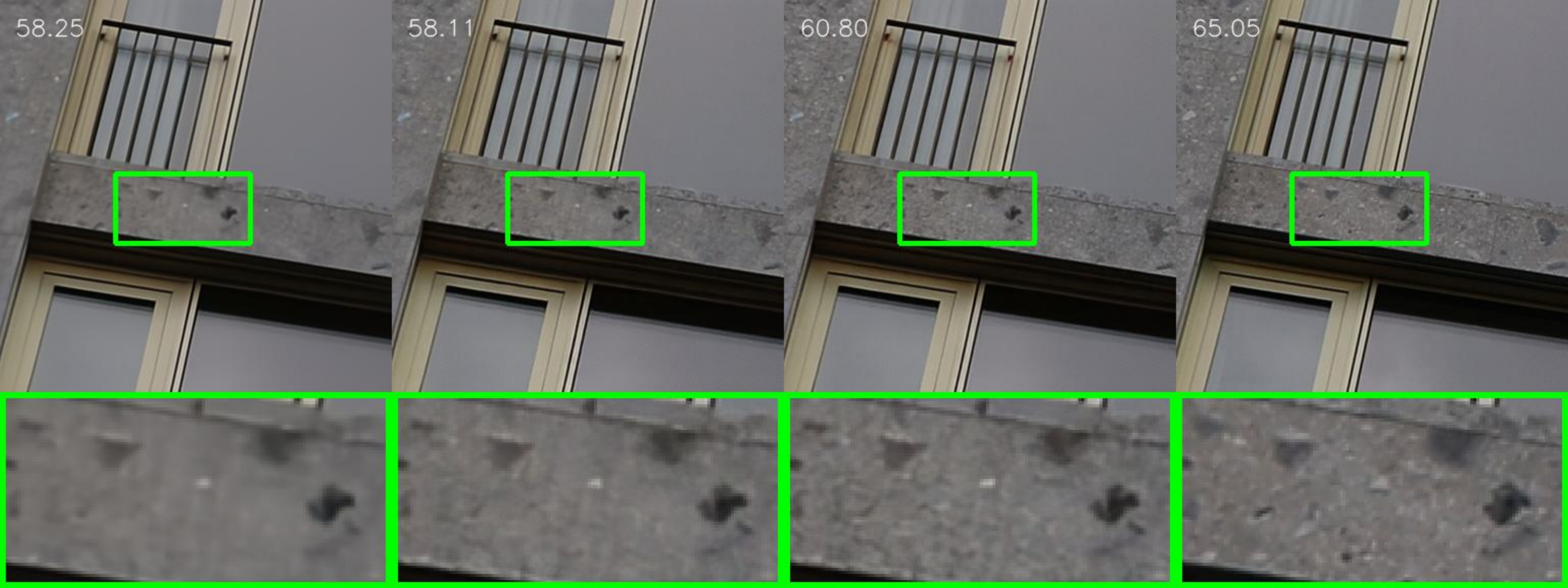}
    \includegraphics[width=0.995\linewidth]{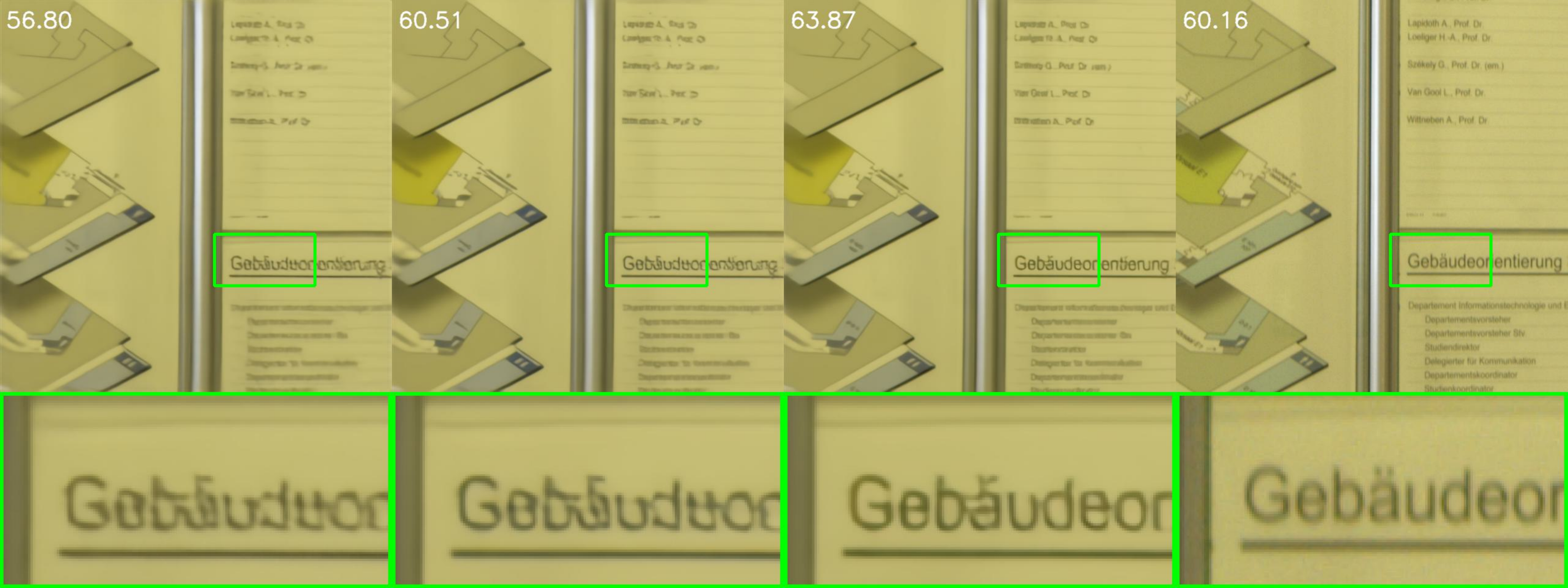}
    \includegraphics[width=0.995\linewidth]{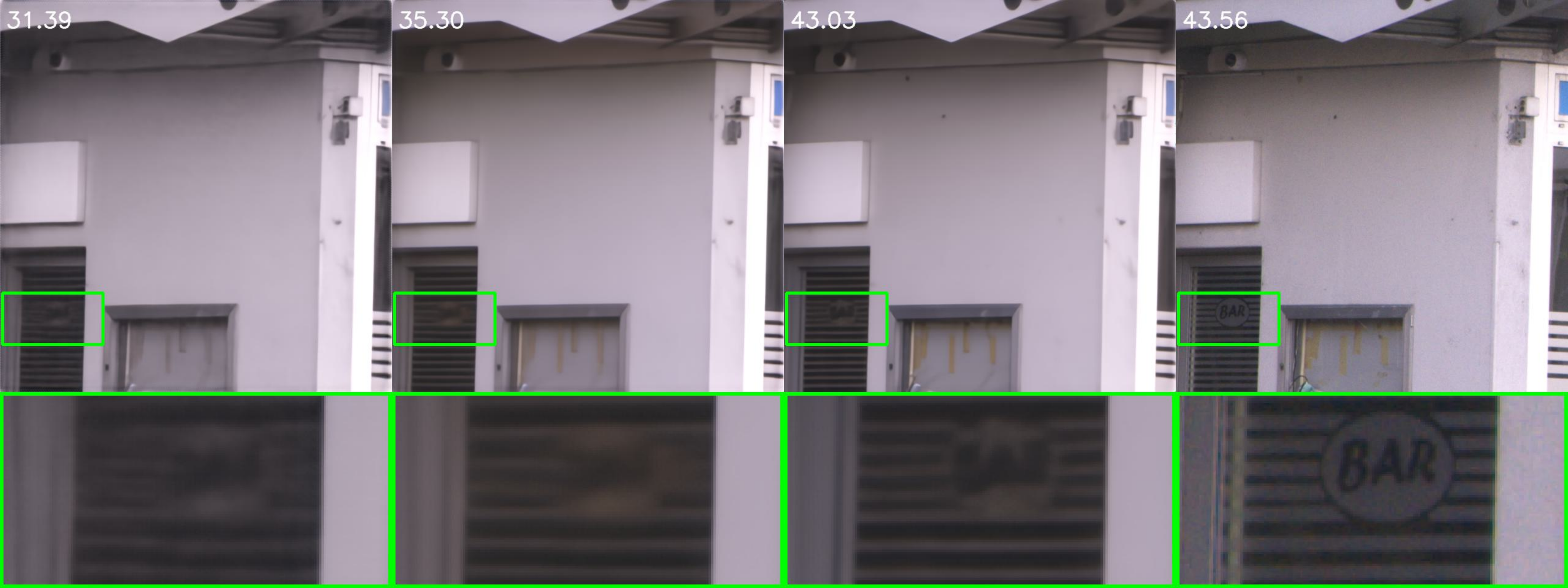}
        \begin{minipage}{0.249\linewidth}
        \centering
        BSRT-L
    \end{minipage}\hfill%
    \begin{minipage}{0.249\linewidth}
        \centering
        BurstGP-BSRT-L \\
        Single-Frame
    \end{minipage}\hfill%
    \begin{minipage}{0.249\linewidth}
        \centering
        BurstGP-BSRT-L \\
        Multiframe (ours)
    \end{minipage}\hfill%
    \begin{minipage}{0.249\linewidth}
        \centering
        GT
    \end{minipage}
    \caption{
        Visualization of the qualitative effects of the single-frame diffusion model, compared to our full multiframe model, on SyntheticBurst (top row) and BurstSR (bottom two rows).  
        We observe that access to the full burst (post-BISR processing) allows the multiframe model to repair certain defects in the single-frame version, such as oversmoothing (first row) or incorrect image content (last two rows). 
    }
    \label{fig:qual-single-abl}
\end{figure}

Further, in Fig.~\ref{fig:qual-single-abl}, we visualize the outputs of the  ``single-frame diffusion'' ablation of our model, by sending only the restored reference from the base BISR model to the generator.
The multiframe form of BurstGP uses the additional data (frames) to either
infer additional high-frequency texture (first row) or 
to resolve additional details (last two rows).

\begin{table}[t]
	\caption{Quantitative evaluations of timestep $t$ on SyntheticBurst dataset.}
	\footnotesize
	\centering
	\begin{adjustbox}{width=0.83\linewidth}
\begin{tabular}{cccccccc}
\toprule
Method         & PSNR$~\uparrow$ & PSNR-L$~\uparrow$ & SSIM$~\uparrow$ & SSIM-L$~\uparrow$ & LPIPS$~\downarrow$ & TOPIQ$~\uparrow$ & MUSIQ$~\uparrow$\\ 
\midrule
$t = 349$ & 26.43 & 35.03 & 0.8941 & 0.9100 & 0.2611 & 0.7808 & 46.10 \\ 
$t = 390$ & 28.18 & 36.44 & 0.8637 & 0.9252 & 0.2504 & 0.7974 & 45.53 \\ 
$t = 399$ & \underline{34.18} & \underline{41.01} & \underline{0.9189} & \underline{0.9600} & \underline{0.1557} & \underline{0.8906} & \underline{46.64} \\ 
$t = 449$ & 23.17 & 32.14 & 0.7614 & 0.8460 & 0.3613 & 0.6294 & 44.85 \\ 
\rowcolor{blue!15} Ours & \textbf{34.29} &   \textbf{41.08}  & \textbf{0.9194} & \textbf{0.9602} & \textbf{0.1543}     &  \textbf{0.8913}   &   \textbf{46.72}  \\ 
\bottomrule
\end{tabular}
	\end{adjustbox}
	\label{tab:syn_timestep}
\end{table}

\section{RealBSR-RAW: Additional Details and Results}
\label{supp:realbsr}

\subsection{Implementation Details}

\noindent
\textbf{FBANet}.
Using the partial implementation released by the authors\footnote{Please see the ``Issues'' of the Github Repository for the FBANet paper, where the authors discuss the implementation on the raw dataset, specifically issues 10 and 13.}
we train FBANet~\cite{wei2023towards} on RealBSR-RAW.
Architecturally, the RGB-space model is reused by applying an initial demosaicing algorithm \cite{menon2006demosaicing}.
We train for 200 epochs with AdamW \cite{loshchilov2017decoupled}, using a batch size of eight bursts per GPU, on four GPUs.

In terms of loss, following the FBANet paper and code, we combine a Charbonnier loss (a robust form of the $L_1$ loss) with the contextual bilateral (CoBi) loss \cite{zhang2019zoom}, via
$
\mathcal{L}_{\mathrm{FBA}} = \mathcal{L}_{\mathrm{Charbonnier}}(I_{\mathrm{GT}}, I_{\mathrm{SR}}) + w_\mathrm{CoBi} \mathcal{L}_{\mathrm{CoBi}}, 
$
where we set $ w_\mathrm{CoBi} = 0.5 $. 
Following Zhang \etal~\cite{zhang2019zoom}, to implement the CoBi loss, we combine an RGB patch loss with a VGG \cite{simonyan2014very} feature loss:
\begin{equation}
    \mathcal{L}_{\mathrm{CoBi}}
    =
    \mathrm{CoBi}_{\mathrm{RGB}}(I_{\mathrm{GT}}, I_{\mathrm{SR}})
    +
    \gamma_{\mathrm{VGG}}
    \mathrm{CoBi}_{\mathrm{VGG}}(I_{\mathrm{GT}}, I_{\mathrm{SR}})
\end{equation}
The RGB patch loss (first term) uses $10\times 10$ patches. Both terms use cosine distance to measure feature dissimilarity.
We set the spatial awareness weight of each CoBi term to $w_s = 0.1$ (using the notation of \cite{zhang2019zoom}) and the VGG weight to $\gamma_{\mathrm{VGG}} = 0.1$, based on its larger natural magnitude.

\noindent
\textbf{BurstM}. 
We initialize BurstM~\cite{kang2024burstm} with the official checkpoint trained on the SyntheticBurst training set. We then train for 200 epochs with the $L_1$ loss in raw space (following the unaligned protocol of FBANet on RealBSR-RAW). We use an initial learning rate of $10^{-5}$, cosine-annealed to $10^{-6}$, and a batch size of ten bursts per GPU, on four GPUs. Other settings follow the default BurstM training setup. We select the checkpoint of the best epoch based on a small held-out validation set from the training data. 
Note that we did attempt to train BurstM with CoBi as well; however, the resulting model was inferior across all metrics. 
We conjecture this may be due to the learned flow model and internal warping procedure interacting with the nonlocal loss.
We leave further investigation to future work.

\subsection{Qualitative Results}

\begin{figure}
    \centering
    \includegraphics[width=0.99\linewidth]{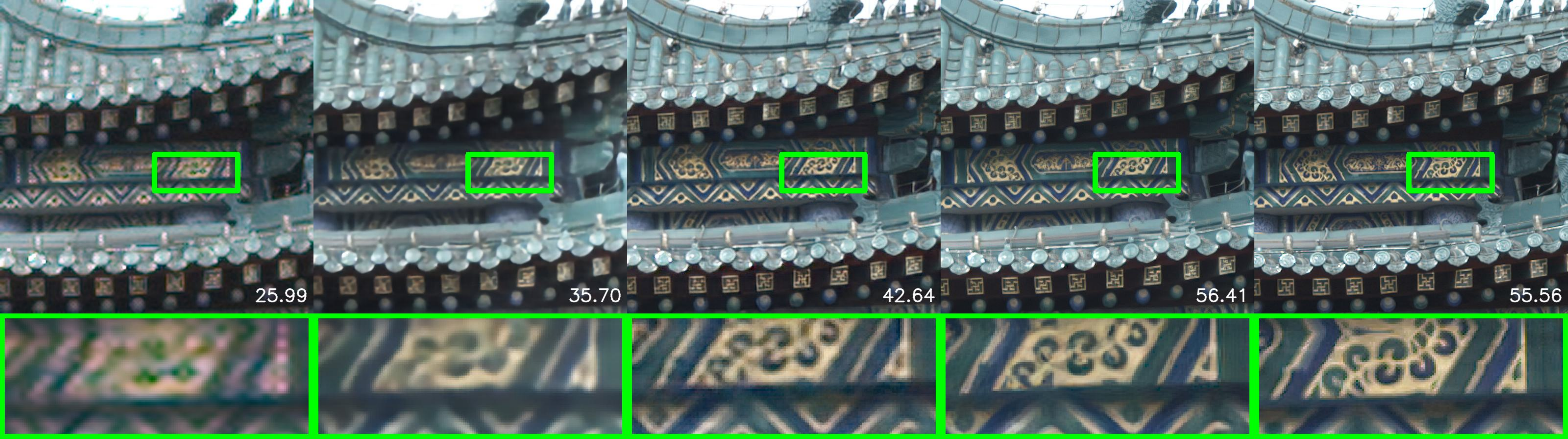}
    \includegraphics[width=0.99\linewidth]{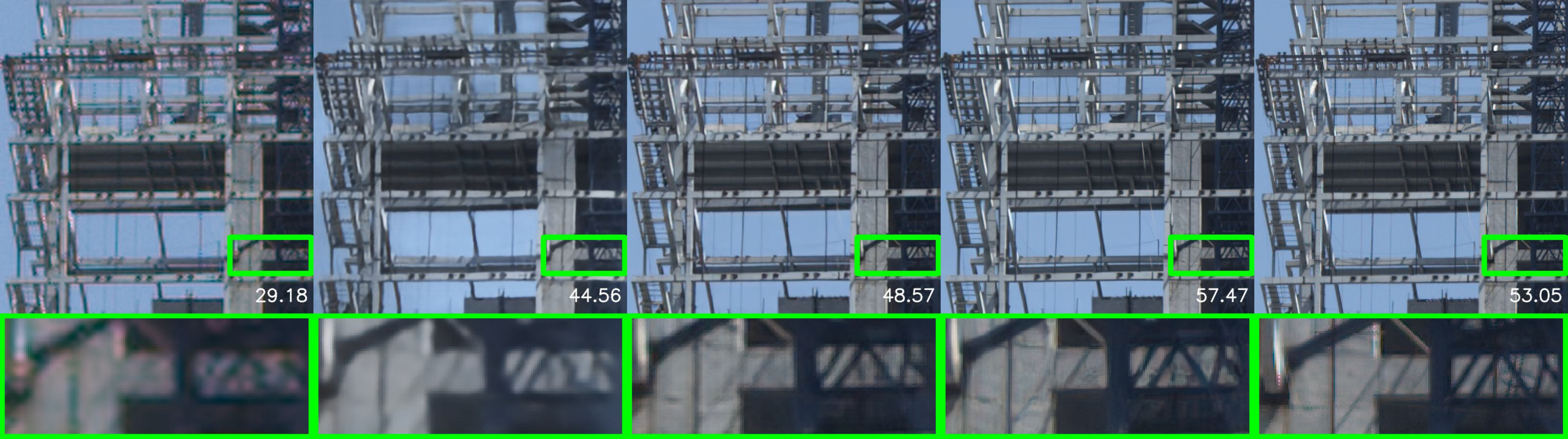}
    \includegraphics[width=0.99\linewidth]{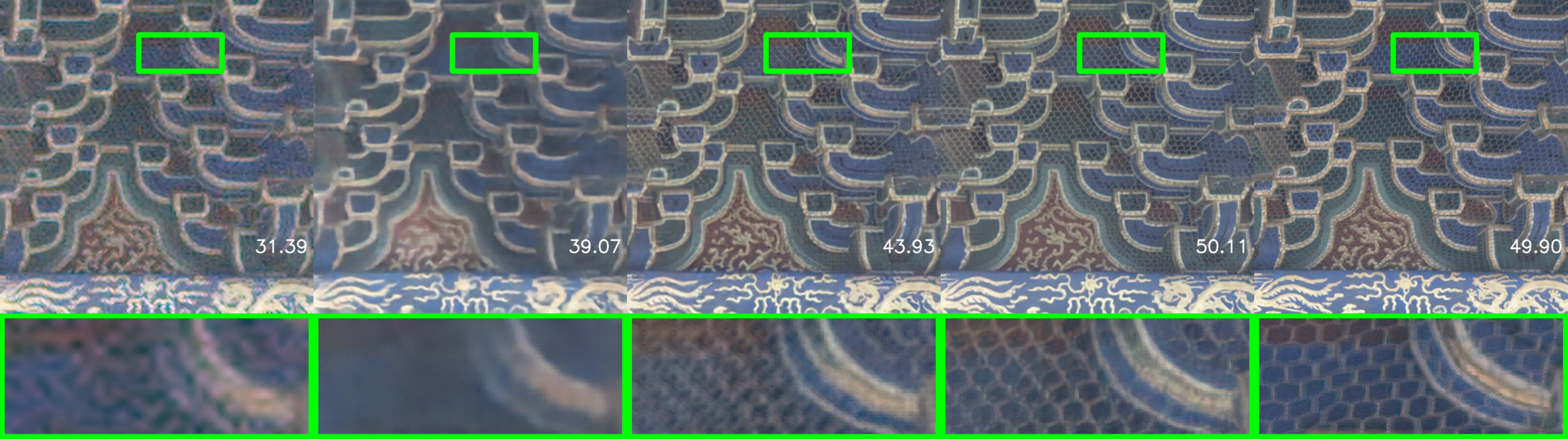}
    \includegraphics[width=0.99\linewidth]{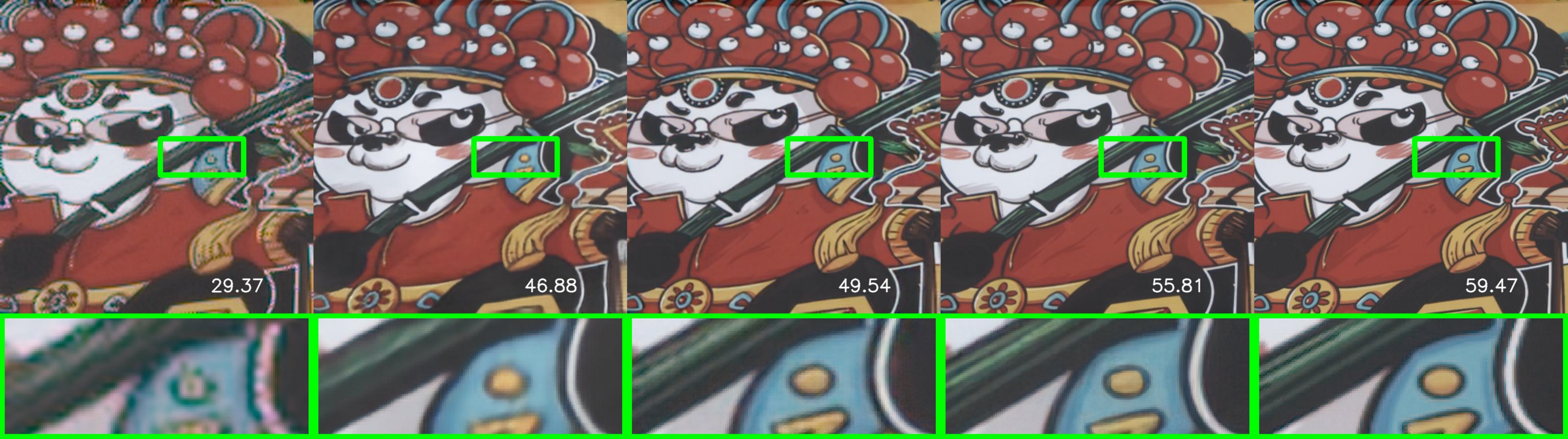}
    \begin{minipage}{0.195\linewidth}
        \centering
        ~Bicubic
    \end{minipage}\hfill%
    \begin{minipage}{0.195\linewidth}
        \centering
        BurstM
    \end{minipage}\hfill%
    \begin{minipage}{0.195\linewidth}
        \centering
        FBANet
    \end{minipage}\hfill%
    \begin{minipage}{0.195\linewidth}
        \centering
        BurstGP-- FBANet (Ours)
    \end{minipage}\hfill%
    \begin{minipage}{0.195\linewidth}
        \centering
        GT~
    \end{minipage}%
    \caption{
    Qualitative Examples from the RealBSR-RAW test set of real bursts.
    From left to right:
    bicubic, BurstM, FBANet, BurstGP-FBANet (ours), and the GT.
    We observe that our method is able to sharpen and deblur the FBANet output, without excessive hallucinations (rows one and two).
    In some cases, it is able to repair damaged image structure, such as the periodic mesh structure in row three, while in others it has fixed certain photographic artifacts 
    (see the final row, where our method has denoised the green object and removed the colour fringing, compared to FBANet).
    }
    \label{fig:realbsr:qual}
\end{figure}

In Fig.~\ref{fig:realbsr:qual}, we display qualitative examples from our model on the RealBSR-RAW \cite{wei2023towards} test set.
Compared to FBANet~\cite{wei2023towards}, upon which our model builds, we observe 
(i) sharper, cleaner outputs (e.g., in row one, there is reduced noise in flat regions);
(ii) additional details without excessive hallucinations (e.g., in row two, we see minor texture added to the stone, slightly enhancing image quality);
(iii) recovery of damaged image structure (e.g., in row three, we observe a striking case of this, with the net-like structure); and
(iv) photographic artifact reduction (e.g., in row four, our model repairs the colour fringing of the FBANet output, at the black-white edge in the zoomed inset).
%
Overall, our approach is not only able to improve perceptual image quality, via the use of a generative model, but can also repair artifacts or errors that persist through the base BISR model in many cases.

\clearpage  


%
%
\bibliographystyle{splncs04}
\bibliography{main}

@String(CVPR  = {IEEE Conf. Comput. Vis. Pattern Recog.})

@String(ICCV  = {Int. Conf. Comput. Vis.})

@String(ECCV  = {Eur. Conf. Comput. Vis.})

@String(NeurIPS = {Adv. Neural Inform. Process. Syst.})

@String(ICLR  = {Int. Conf. Learn. Represent.})

@String(CVPRW = {IEEE Conf. Comput. Vis. Pattern Recog. Worksh.})

@String(TOG   = {ACM Trans. Graph.})

@String(TIP   = {IEEE Trans. Image Process.})

@String(CVPR  = {CVPR})

@String(ICCV  = {ICCV})

@String(ECCV  = {ECCV})

@String(NeurIPS = {NeurIPS})

@String(ICLR  = {ICLR})

@String(CVPRW = {CVPRW})

@String(TOG   = {ACM TOG})

@String(TIP   = {IEEE TIP})

@String { AI           = {{Artificial Intelligence}} }

@String { CVPR         = {{Proceedings of the {IEEE} Conference on Computer Vision and Pattern Recognition ({CVPR})}} }

@String { CVPRW        = {{Proceedings of the {IEEE} Conference on Computer Vision and Pattern Recognition Workshops ({CVPRW})}} }

@String { ECCV         = {{Proceedings of the European Conference on Computer Vision ({ECCV})}} }

@String { ICCV         = {{Proceedings of the International Conference on Computer Vision ({ICCV})}} }

@String { ICLR         = {{Proceedings of the International Conference on Learning Representations ({ICLR})}} }

@String { IJCNN        = {{Proceedings of the International Joint Conference on Neural Networks ({IJCNN})}} }

@String { NeurIPS      = {{Neural Information Processing Systems ({NeurIPS})}} }

@String { TIP          = {{{IEEE} Transactions on Image Processing}} }

@String { TOG          = {{{{ACM} Transactions on Graphics ({TOG})}}} }

@String { CVPR         = {{{IEEE} Conference on Computer Vision and Pattern Recognition ({CVPR})}} }

@String { CVPRW        = {{{IEEE} Conference on Computer Vision and Pattern Recognition Workshops ({CVPRW})}} }

@String { ECCV         = {{European Conference on Computer Vision ({ECCV})}} }

@String { ICCV         = {{International Conference on Computer Vision ({ICCV})}} }

@String { ICLR         = {{International Conference on Learning Representations ({ICLR})}} }

@inproceedings{bhat2021deep,
  title={Deep burst super-resolution},
  author={Bhat, Goutam and Danelljan, Martin and Van Gool, Luc and Timofte, Radu},
  booktitle=CVPR,
  pages={9209--9218},
  year={2021}
}

@inproceedings{DosovitskiyB0WZ21,
  author       = {Alexey Dosovitskiy and
                  Lucas Beyer and
                  Alexander Kolesnikov and
                  Dirk Weissenborn and
                  Xiaohua Zhai and
                  Thomas Unterthiner and
                  Mostafa Dehghani and
                  Matthias Minderer and
                  Georg Heigold and
                  Sylvain Gelly and
                  Jakob Uszkoreit and
                  Neil Houlsby},
  title        = {An Image is Worth 16x16 Words: Transformers for Image Recognition
                  at Scale},
  booktitle    = ICLR,
  year         = {2021}
}

@article{LeCunBBH98,
  author       = {Yann LeCun and
                  L{\'{e}}on Bottou and
                  Yoshua Bengio and
                  Patrick Haffner},
  title        = {Gradient-based learning applied to document recognition},
  journal      = {Proc. {IEEE}},
  volume       = {86},
  number       = {11},
  pages        = {2278--2324},
  year         = {1998}
}

@inproceedings{dudhane2023burstormer,
  title={Burstormer: Burst image restoration and enhancement transformer},
  author={Dudhane, Akshay and Zamir, Syed Waqas and Khan, Salman and Khan, Fahad Shahbaz and Yang, Ming-Hsuan},
  booktitle=CVPR,
  pages={5703--5712},
  year={2023}
}

@inproceedings{kang2024burstm,
  title={{BurstM}: Deep burst multi-scale {SR} using {F}ourier space with optical flow},
  author={Kang, EungGu and Lee, Byeonghun and Im, Sunghoon and Jin, Kyong Hwan},
  booktitle=ECCV,
  pages={459--477},
  year={2024}
}

@inproceedings{dai2007soft,
  title={Soft edge smoothness prior for alpha channel super resolution},
  author={Dai, Shengyang and Han, Mei and Xu, Wei and Wu, Ying and Gong, Yihong},
  booktitle=CVPR,
  year={2007},
}

@article{li2001new,
  title={New edge-directed interpolation},
  author={Li, Xin and Orchard, Michael T},
  journal=TIP,
  volume={10},
  number={10},
  pages={1521--1527},
  year={2001}
}

@inproceedings{bhat2021deep2,
  title={Deep reparametrization of multi-frame super-resolution and denoising},
  author={Bhat, Goutam and Danelljan, Martin and Yu, Fisher and Van Gool, Luc and Timofte, Radu},
  booktitle=ICCV,
  pages={2460--2470},
  year={2021}
}

@inproceedings{dudhane2022burst,
  title={Burst image restoration and enhancement},
  author={Dudhane, Akshay and Zamir, Syed Waqas and Khan, Salman and Khan, Fahad Shahbaz and Yang, Ming-Hsuan},
  booktitle=CVPR,
  pages={5759--5768},
  year={2022}
}

@inproceedings{luo2022bsrt,
  title={{BSRT}: Improving burst super-resolution with {Swin} transformer and flow-guided deformable alignment},
  author={Luo, Ziwei and Li, Youwei and Cheng, Shen and Yu, Lei and Wu, Qi and Wen, Zhihong and Fan, Haoqiang and Sun, Jian and Liu, Shuaicheng},
  booktitle=CVPR,
  pages={998--1008},
  year={2022}
}

@inproceedings{wei2023towards,
  title={Towards real-world burst image super-resolution: Benchmark and method},
  author={Wei, Pengxu and Sun, Yujing and Guo, Xingbei and Liu, Chang and Li, Guanbin and Chen, Jie and Ji, Xiangyang and Lin, Liang},
  booktitle=ICCV,
  pages={13233--13242},
  year={2023}
}

@inproceedings{luo2021ebsr,
  title={{EBSR}: Feature enhanced burst super-resolution with deformable alignment},
  author={Luo, Ziwei and Yu, Lei and Mo, Xuan and Li, Youwei and Jia, Lanpeng and Fan, Haoqiang and Sun, Jian and Liu, Shuaicheng},
  booktitle=CVPR,
  pages={471--478},
  year={2021}
}

@article{ho2020denoising,
  title={Denoising diffusion probabilistic models},
  author={Ho, Jonathan and Jain, Ajay and Abbeel, Pieter},
  journal=NeurIPS,
  volume={33},
  pages={6840--6851},
  year={2020}
}

@inproceedings{rombach2022high,
  title={High-resolution image synthesis with latent diffusion models},
  author={Rombach, Robin and Blattmann, Andreas and Lorenz, Dominik and Esser, Patrick and Ommer, Bj{\"o}rn},
  booktitle=CVPR,
  pages={10684--10695},
  year={2022}
}

@inproceedings{peebles2023scalable,
  title={Scalable diffusion models with transformers},
  author={Peebles, William and Xie, Saining},
  booktitle=ICCV,
  pages={4195--4205},
  year={2023}
}

@article{podell2023sdxl,
  title={{SDXL}: Improving latent diffusion models for high-resolution image synthesis},
  author={Podell, Dustin and English, Zion and Lacey, Kyle and Blattmann, Andreas and Dockhorn, Tim and M{\"u}ller, Jonas and Penna, Joe and Rombach, Robin},
  journal={arXiv preprint arXiv:2307.01952},
  year={2023}
}

@article{labs2025flux,
  title={{FLUX.1} {Kontext}: Flow Matching for In-Context Image Generation and Editing in Latent Space},
  author={Labs, Black Forest and Batifol, Stephen and Blattmann, Andreas and Boesel, Frederic and Consul, Saksham and Diagne, Cyril and Dockhorn, Tim and English, Jack and English, Zion and Esser, Patrick and others},
  journal={arXiv preprint arXiv:2506.15742},
  year={2025}
}

@inproceedings{lin2024diffbir,
  title={{DiffBIR}: Toward blind image restoration with generative diffusion prior},
  author={Lin, Xinqi and He, Jingwen and Chen, Ziyan and Lyu, Zhaoyang and Dai, Bo and Yu, Fanghua and Qiao, Yu and Ouyang, Wanli and Dong, Chao},
  booktitle=ECCV,
  pages={430--448},
  year={2024}
}

@article{wang2024exploiting,
  title={Exploiting diffusion prior for real-world image super-resolution},
  author={Wang, Jianyi and Yue, Zongsheng and Zhou, Shangchen and Chan, Kelvin CK and Loy, Chen Change},
  journal={International Journal of Computer Vision},
  volume={132},
  number={12},
  pages={5929--5949},
  year={2024},
  publisher={Springer}
}

@inproceedings{sun2025pixel,
  title={Pixel-level and semantic-level adjustable super-resolution: A dual-lora approach},
  author={Sun, Lingchen and Wu, Rongyuan and Ma, Zhiyuan and Liu, Shuaizheng and Yi, Qiaosi and Zhang, Lei},
  booktitle=CVPR,
  pages={2333--2343},
  year={2025}
}

@inproceedings{wang2025seedvr,
  title={{SeedVR}: Seeding infinity in diffusion transformer towards generic video restoration},
  author={Wang, Jianyi and Lin, Zhijie and Wei, Meng and Zhao, Yang and Yang, Ceyuan and Loy, Chen Change and Jiang, Lu},
  booktitle=CVPR,
  pages={2161--2172},
  year={2025}
}

@inproceedings{xie2025star,
  title={{STAR}: Spatial-temporal augmentation with text-to-video models for real-world video super-resolution},
  author={Xie, Rui and Liu, Yinhong and Zhou, Penghao and Zhao, Chen and Zhou, Jun and Zhang, Kai and Zhang, Zhenyu and Yang, Jian and Yang, Zhenheng and Tai, Ying},
  booktitle=ICCV,
  pages={17108--17118},
  year={2025}
}

@article{chen2025dove,
  title={{DOVE}: Efficient one-step diffusion model for real-world video super-resolution},
  author={Chen, Zheng and Zou, Zichen and Zhang, Kewei and Su, Xiongfei and Yuan, Xin and Guo, Yong and Zhang, Yulun},
  journal=NeurIPS,
  year={2025}
}

@inproceedings{yi2025fine,
  title={Fine-structure preserved real-world image super-resolution via transfer vae training},
  author={Yi, Qiaosi and Li, Shuai and Wu, Rongyuan and Sun, Lingchen and Wu, Yuhui and Zhang, Lei},
  booktitle=ICCV,
  pages={12415--12426},
  year={2025}
}

@article{kim2025chain,
  title={Chain-of-zoom: Extreme super-resolution via scale autoregression and preference alignment},
  author={Kim, Bryan Sangwoo and Kim, Jeongsol and Ye, Jong Chul},
  journal=NEURIPS,
  year={2025}
}

@inproceedings{chen2025faithdiff,
  title={{FaithDiff}: Unleashing diffusion priors for faithful image super-resolution},
  author={Chen, Junyang and Pan, Jinshan and Dong, Jiangxin},
  booktitle=CVPR,
  pages={28188--28197},
  year={2025}
}

@article{huang2025diffusion,
  title={Diffusion model-based image editing: A survey},
  author={Huang, Yi and Huang, Jiancheng and Liu, Yifan and Yan, Mingfu and Lv, Jiaxi and Liu, Jianzhuang and Xiong, Wei and Zhang, He and Cao, Liangliang and Chen, Shifeng},
  journal={IEEE Transactions on Pattern Analysis and Machine Intelligence},
  year={2025},
  publisher={IEEE}
}

@inproceedings{blau2018perception,
  title={The perception-distortion tradeoff},
  author={Blau, Yochai and Michaeli, Tomer},
  booktitle=CVPR,
  pages={6228--6237},
  year={2018}
}

@article{baldridge2024imagen,
  title={Imagen 3},
  author={Baldridge, Jason and Bauer, Jakob and Bhutani, Mukul and Brichtova, Nicole and Bunner, Andrew and Castrejon, Lluis and Chan, Kelvin and Chen, Yichang and Dieleman, Sander and Du, Yuqing and others},
  journal={arXiv preprint arXiv:2408.07009},
  year={2024}
}

@inproceedings{wei2025perceive,
  title={Perceive, understand and restore: Real-world image super-resolution with autoregressive multimodal generative models},
  author={Wei, Hongyang and Liu, Shuaizheng and Yuan, Chun and Zhang, Lei},
  booktitle=ICCV,
  pages={18640--18650},
  year={2025}
}

@inproceedings{li2022face,
  title={From face to natural image: Learning real degradation for blind image super-resolution},
  author={Li, Xiaoming and Chen, Chaofeng and Lin, Xianhui and Zuo, Wangmeng and Zhang, Lei},
  booktitle=ECCV,
  pages={376--392},
  year={2022},
  organization={Springer}
}

@article{chen2024restoreagent,
  title={{RestoreAgent}: Autonomous image restoration agent via multimodal large language models},
  author={Chen, Haoyu and Li, Wenbo and Gu, Jinjin and Ren, Jingjing and Chen, Sixiang and Ye, Tian and Pei, Renjing and Zhou, Kaiwen and Song, Fenglong and Zhu, Lei},
  journal=NeurIPS,
  volume={37},
  pages={110643--110666},
  year={2024}
}

@inproceedings{duan2025dit4sr,
  title={{DiT4SR}: Taming diffusion transformer for real-world image super-resolution},
  author={Duan, Zheng-Peng and Zhang, Jiawei and Jin, Xin and Zhang, Ziheng and Xiong, Zheng and Zou, Dongqing and Ren, Jimmy S and Guo, Chunle and Li, Chongyi},
  booktitle=ICCV,
  pages={18948--18958},
  year={2025}
}

@article{zhang2024degradation,
  title={Degradation-guided one-step image super-resolution with diffusion priors},
  author={Zhang, Aiping and Yue, Zongsheng and Pei, Renjing and Ren, Wenqi and Cao, Xiaochun},
  journal={arXiv preprint arXiv:2409.17058},
  year={2024}
}

@inproceedings{kong2025dam,
  title={{DAM-VSR}: Disentanglement of appearance and motion for video super-resolution},
  author={Kong, Zhe and Li, Le and Zhang, Yong and Gao, Feng and Yang, Shaoshu and Wang, Tao and Zhang, Kaihao and Kang, Zhuoliang and Wei, Xiaoming and Chen, Guanying and Luo, Wenhan},
  booktitle={Proceedings of the Special Interest Group on Computer Graphics and Interactive Techniques Conference Conference Papers},
  pages={1--11},
  year={2025}
}

@article{zhao2024avernet,
  title={{AverNet}: All-in-one video restoration for time-varying unknown degradations},
  author={Zhao, Haiyu and Tian, Lei and Xiao, Xinyan and Hu, Peng and Gou, Yuanbiao and Peng, Xi},
  journal={Advances in Neural Information Processing Systems},
  volume={37},
  pages={127296--127316},
  year={2024}
}

@inproceedings{chan2022investigating,
  title={Investigating tradeoffs in real-world video super-resolution},
  author={Chan, Kelvin CK and Zhou, Shangchen and Xu, Xiangyu and Loy, Chen Change},
  booktitle={Proceedings of the IEEE/CVF conference on computer vision and pattern recognition},
  pages={5962--5971},
  year={2022}
}

@inproceedings{mao2025making,
  title={Making old film great again: Degradation-aware state space model for old film restoration},
  author={Mao, Yudong and Luo, Hao and Zhong, Zhiwei and Chen, Peilin and Zhang, Zhijiang and Wang, Shiqi},
  booktitle=CVPR,
  pages={28039--28049},
  year={2025}
}

@inproceedings{yang2024motion,
  title={Motion-guided latent diffusion for temporally consistent real-world video super-resolution},
  author={Yang, Xi and He, Chenhang and Ma, Jianqi and Zhang, Lei},
  booktitle=ECCV,
  pages={224--242},
  year={2024},
  organization={Springer}
}

@inproceedings{zhang2024realviformer,
  title={{RealViformer}: Investigating attention for real-world video super-resolution},
  author={Zhang, Yuehan and Yao, Angela},
  booktitle=ECCV,
  pages={412--428},
  year={2024},
  organization={Springer}
}

@article{shi2022rethinking,
  title={Rethinking alignment in video super-resolution transformers},
  author={Shi, Shuwei and Gu, Jinjin and Xie, Liangbin and Wang, Xintao and Yang, Yujiu and Dong, Chao},
  journal={Advances in Neural Information Processing Systems},
  volume={35},
  pages={36081--36093},
  year={2022}
}

@inproceedings{wang2025turbovsr,
  title={{TurboVSR}: Fantastic video upscalers and where to find them},
  author={Wang, Zhongdao and Zhao, Guodongfang and Ren, Jingjing and Feng, Bailan and Zhang, Shifeng and Li, Wenbo},
  booktitle=ICCV,
  pages={18132--18142},
  year={2025}
}

@article{ren2025hallucination,
  title={Hallucination Score: Towards Mitigating Hallucinations in Generative Image Super-Resolution},
  author={Ren, Weiming and Goyal, Raghav and Hu, Zhiming and Aumentado-Armstrong, Tristan Ty and Mohomed, Iqbal and Levinshtein, Alex},
  journal={arXiv preprint arXiv:2507.14367},
  year={2025}
}

@inproceedings{yu2024scaling,
  title={Scaling up to excellence: Practicing model scaling for photo-realistic image restoration in the wild},
  author={Yu, Fanghua and Gu, Jinjin and Li, Zheyuan and Hu, Jinfan and Kong, Xiangtao and Wang, Xintao and He, Jingwen and Qiao, Yu and Dong, Chao},
  booktitle=CVPR,
  pages={25669--25680},
  year={2024}
}

@inproceedings{wu2024seesr,
  title={{SeeSR}: Towards semantics-aware real-world image super-resolution},
  author={Wu, Rongyuan and Yang, Tao and Sun, Lingchen and Zhang, Zhengqiang and Li, Shuai and Zhang, Lei},
  booktitle={Proceedings of the IEEE/CVF conference on computer vision and pattern recognition},
  pages={25456--25467},
  year={2024}
}

@article{wu2024one,
  title={One-step effective diffusion network for real-world image super-resolution},
  author={Wu, Rongyuan and Sun, Lingchen and Ma, Zhiyuan and Zhang, Lei},
  journal={Advances in Neural Information Processing Systems},
  volume={37},
  pages={92529--92553},
  year={2024}
}

@inproceedings{ghildyal2022shift,
  title={Shift-tolerant perceptual similarity metric},
  author={Ghildyal, Abhijay and Liu, Feng},
  booktitle=ECCV,
  year={2022},
}

@inproceedings{noroozi2024you,
  title={You only need one step: Fast super-resolution with stable diffusion via scale distillation},
  author={Noroozi, Mehdi and Hadji, Isma and Martinez, Brais and Bulat, Adrian and Tzimiropoulos, Georgios},
  booktitle=ECCV,
  pages={145--161},
  year={2024},
  organization={Springer}
}

@inproceedings{kawai2025efficient,
  title={Efficient Burst Super-Resolution with One-step Diffusion},
  author={Kawai, Kento and Oba, Takeru and Tokoro, Kyotaro and Akita, Kazutoshi and Ukita, Norimichi},
  booktitle=CVPR,
  pages={864--873},
  year={2025}
}

@article{moser2024zoomed,
  title={Zoomed in, diffused out: Towards local degradation-aware multi-diffusion for extreme image super-resolution},
  author={Moser, Brian B and Frolov, Stanislav and Nauen, Tobias C and Raue, Federico and Dengel, Andreas},
  journal={arXiv preprint arXiv:2411.12072},
  year={2024}
}

@inproceedings{lipmanflow,
  title={Flow Matching for Generative Modeling},
  author={Lipman, Yaron and Chen, Ricky TQ and Ben-Hamu, Heli and Nickel, Maximilian and Le, Matthew},
  booktitle=ICLR,
  year={2023}
}

@article{zhuang2025flashvsr,
  title={{FlashVSR}: Towards real-time diffusion-based streaming video super-resolution},
  author={Zhuang, Junhao and Guo, Shi and Cai, Xin and Li, Xiaohui and Liu, Yihao and Yuan, Chun and Xue, Tianfan},
  journal={arXiv preprint arXiv:2510.12747},
  year={2025}
}

@article{xie2025simplegvr,
  title={{SimpleGVR}: A simple baseline for latent-cascaded video super-resolution},
  author={Xie, Liangbin and Li, Yu and Du, Shian and Xia, Menghan and Wang, Xintao and Yu, Fanghua and Chen, Ziyan and Wan, Pengfei and Zhou, Jiantao and Dong, Chao},
  journal=ICLR,
  year={2026}
}

@inproceedings{sun2025one,
  title={One-step diffusion for detail-rich and temporally consistent video super-resolution},
  author={Sun, Yujing and Sun, Lingchen and Liu, Shuaizheng and Wu, Rongyuan and Zhang, Zhengqiang and Zhang, Lei},
  booktitle=NeurIPS,
  year={2025}
}

@inproceedings{zhou2024upscale,
  title={Upscale-a-video: Temporal-consistent diffusion model for real-world video super-resolution},
  author={Zhou, Shangchen and Yang, Peiqing and Wang, Jianyi and Luo, Yihang and Loy, Chen Change},
  booktitle=CVPR,
  pages={2535--2545},
  year={2024}
}

@article{yang2024cogvideox,
  title={{CogVideoX}: Text-to-video diffusion models with an expert transformer},
  author={Yang, Zhuoyi and Teng, Jiayan and Zheng, Wendi and Ding, Ming and Huang, Shiyu and Xu, Jiazheng and Yang, Yuanming and Hong, Wenyi and Zhang, Xiaohan and Feng, Guanyu and others},
  journal={arXiv preprint arXiv:2408.06072},
  year={2024}
}

@article{blattmann2023stable,
  title={Stable video diffusion: Scaling latent video diffusion models to large datasets},
  author={Blattmann, Andreas and Dockhorn, Tim and Kulal, Sumith and Mendelevitch, Daniel and Kilian, Maciej and Lorenz, Dominik and Levi, Yam and English, Zion and Voleti, Vikram and Letts, Adam and others},
  journal={arXiv preprint arXiv:2311.15127},
  year={2023}
}

@article{irani1991improving,
  title={Improving resolution by image registration},
  author={Irani, Michal and Peleg, Shmuel},
  journal={CVGIP: Graphical models and image processing},
  volume={53},
  number={3},
  pages={231--239},
  year={1991},
  publisher={Elsevier}
}

@article{hasinoff2016burst,
  title={Burst photography for high dynamic range and low-light imaging on mobile cameras},
  author={Hasinoff, Samuel W and Sharlet, Dillon and Geiss, Ryan and Adams, Andrew and Barron, Jonathan T and Kainz, Florian and Chen, Jiawen and Levoy, Marc},
  journal={ACM Transactions on Graphics (ToG)},
  volume={35},
  number={6},
  pages={1--12},
  year={2016},
  publisher={ACM New York, NY, USA}
}

@inproceedings{liu2025ultravsr,
  title={{UltraVSR}: Achieving ultra-realistic video super-resolution with efficient one-step diffusion space},
  author={Liu, Yong and Pan, Jinshan and Li, Yinchuan and Dong, Qingji and Zhu, Chao and Guo, Yu and Wang, Fei},
  booktitle={Proceedings of the ACM International Conference on Multimedia},
  pages={7785--7794},
  year={2025}
}

@inproceedings{mildenhall2018burst,
  title={Burst denoising with kernel prediction networks},
  author={Mildenhall, Ben and Barron, Jonathan T and Chen, Jiawen and Sharlet, Dillon and Ng, Ren and Carroll, Robert},
  booktitle=CVPR,
  pages={2502--2510},
  year={2018}
}

@inproceedings{di2025qmambabsr,
  title={{QMambaBSR}: Burst image super-resolution with query state space model},
  author={Di, Xin and Peng, Long and Xia, Peizhe and Li, Wenbo and Pei, Renjing and Cao, Yang and Wang, Yang and Zha, Zheng-Jun},
  booktitle=CVPR,
  pages={23080--23090},
  year={2025}
}

@inproceedings{bhat2023self,
  title={Self-supervised burst super-resolution},
  author={Bhat, Goutam and Gharbi, Micha{\"e}l and Chen, Jiawen and Van Gool, Luc and Xia, Zhihao},
  booktitle=ICCV,
  pages={10605--10614},
  year={2023}
}

@inproceedings{tokoro2024burst,
  title={Burst super-resolution with diffusion models for improving perceptual quality},
  author={Tokoro, Kyotaro and Akita, Kazutoshi and Ukita, Norimichi},
  booktitle=IJCNN,
  pages={1--8},
  year={2024}
}

@article{wang2025seedvr2,
  title={{SeedVR2}: One-step video restoration via diffusion adversarial post-training},
  author={Wang, Jianyi and Lin, Shanchuan and Lin, Zhijie and Ren, Yuxi and Wei, Meng and Yue, Zongsheng and Zhou, Shangchen and Chen, Hao and Zhao, Yang and Yang, Ceyuan and others},
  journal={arXiv preprint arXiv:2506.05301},
  year={2025}
}

@article{wan2025,
      title={Wan: Open and Advanced Large-Scale Video Generative Models}, 
      author={Team Wan and Ang Wang and Baole Ai and Bin Wen and Chaojie Mao and Chen-Wei Xie and Di Chen and Feiwu Yu and Haiming Zhao and Jianxiao Yang and Jianyuan Zeng and Jiayu Wang and Jingfeng Zhang and Jingren Zhou and Jinkai Wang and Jixuan Chen and Kai Zhu and Kang Zhao and Keyu Yan and Lianghua Huang and Mengyang Feng and Ningyi Zhang and Pandeng Li and Pingyu Wu and Ruihang Chu and Ruili Feng and Shiwei Zhang and Siyang Sun and Tao Fang and Tianxing Wang and Tianyi Gui and Tingyu Weng and Tong Shen and Wei Lin and Wei Wang and Wei Wang and Wenmeng Zhou and Wente Wang and Wenting Shen and Wenyuan Yu and Xianzhong Shi and Xiaoming Huang and Xin Xu and Yan Kou and Yangyu Lv and Yifei Li and Yijing Liu and Yiming Wang and Yingya Zhang and Yitong Huang and Yong Li and You Wu and Yu Liu and Yulin Pan and Yun Zheng and Yuntao Hong and Yupeng Shi and Yutong Feng and Zeyinzi Jiang and Zhen Han and Zhi-Fan Wu and Ziyu Liu},
      journal = {arXiv preprint arXiv:2503.20314},
      year={2025}
}

@inproceedings{hu2022lora,
  title={{LoRA}: Low-rank adaptation of large language models.},
  author={Edward J. Hu and
                  Yelong Shen and
                  Phillip Wallis and
                  Zeyuan Allen{-}Zhu and
                  Yuanzhi Li and
                  Shean Wang and
                  Lu Wang and
                  Weizhu Chen},
  booktitle=ICLR,
  year={2022}
}

@book{tikhonov_solutions_1977,
  series = {Scripta {Series} in {Mathematics}},
  title = {Solutions of ill-posed problems},
  isbn = {0-470-99124-0},
  publisher = {V. H. Winston \& Sons},
  author = {Tikhonov, Andrey Nikolaevich and Arsenin, Vasiliy Y.},
  year = {1977},
  note = {10239}
}

@article{hansen1987truncated,
  title={The truncated {SVD} as a method for regularization},
  author={Hansen, Per Christian},
  journal={BIT Numerical Mathematics},
  volume={27},
  number={4},
  pages={534--553},
  year={1987},
  publisher={Springer}
}

@article{loshchilov2017fixing,
  title={Fixing weight decay regularization in {Adam}},
  author={Loshchilov, Ilya and Hutter, Frank and others},
  journal={arXiv preprint arXiv:1711.05101},
  volume={5},
  number={5},
  pages={5},
  year={2017}
}

@inproceedings{zhang2018unreasonable,
  title={The unreasonable effectiveness of deep features as a perceptual metric},
  author={Zhang, Richard and Isola, Phillip and Efros, Alexei A and Shechtman, Eli and Wang, Oliver},
  booktitle=CVPR,
  pages={586--595},
  year={2018}
}

@article{simonyan2014very,
  title={Very deep convolutional networks for large-scale image recognition},
  author={Simonyan, Karen and Zisserman, Andrew},
  journal={arXiv preprint arXiv:1409.1556},
  year={2014}
}

@article{loshchilov2017decoupled,
  title={Decoupled weight decay regularization},
  author={Loshchilov, Ilya and Hutter, Frank},
  journal={arXiv preprint arXiv:1711.05101},
  year={2017}
}

@inproceedings{li2024dualdn,
  title={{DualDn}: Dual-domain denoising via differentiable {ISP}},
  author={Li, Ruikang and Wang, Yujin and Chen, Shiqi and Zhang, Fan and Gu, Jinwei and Xue, Tianfan},
  booktitle=ECCV,
  pages={160--177},
  year={2024},
  organization={Springer}
}

@inproceedings{yu2021reconfigisp,
  title={{ReconfigISP}: Reconfigurable camera image processing pipeline},
  author={Yu, Ke and Li, Zexian and Peng, Yue and Loy, Chen Change and Gu, Jinwei},
  booktitle={Proceedings of the IEEE/CVF International Conference on Computer Vision},
  pages={4248--4257},
  year={2021}
}

@article{menon2006demosaicing,
  title={Demosaicing with directional filtering and a posteriori decision},
  author={Menon, Daniele and Andriani, Stefano and Calvagno, Giancarlo},
  journal={IEEE Transactions on Image Processing},
  volume={16},
  number={1},
  pages={132--141},
  year={2007},
  publisher={IEEE}
}

@inproceedings{brooks2019unprocessing,
  title={Unprocessing images for learned raw denoising},
  author={Brooks, Tim and Mildenhall, Ben and Xue, Tianfan and Chen, Jiawen and Sharlet, Dillon and Barron, Jonathan T},
  booktitle=CVPR,
  pages={11036--11045},
  year={2019}
}

@inproceedings{ignatov2020replacing,
  title={Replacing mobile camera {ISP} with a single deep learning model},
  author={Ignatov, Andrey and Van Gool, Luc and Timofte, Radu},
  booktitle=CVPRW,
  pages={536--537},
  year={2020}
}

@inproceedings{zhang2019zoom,
  title={Zoom to learn, learn to zoom},
  author={Zhang, Xuaner and Chen, Qifeng and Ng, Ren and Koltun, Vladlen},
  booktitle=CVPR,
  pages={3762--3770},
  year={2019}
}

@article{chen2024topiq,
  title={{TOPIQ}: A top-down approach from semantics to distortions for image quality assessment},
  author={Chen, Chaofeng and Mo, Jiadi and Hou, Jingwen and Wu, Haoning and Liao, Liang and Sun, Wenxiu and Yan, Qiong and Lin, Weisi},
  journal=TIP,
  volume={33},
  pages={2404--2418},
  year={2024},
  publisher={IEEE}
}

@inproceedings{ke2021musiq,
  title={{MUSIQ}: Multi-scale image quality transformer},
  author={Ke, Junjie and Wang, Qifei and Wang, Yilin and Milanfar, Peyman and Yang, Feng},
  booktitle=ICCV,
  pages={5148--5157},
  year={2021}
}




\end{document}